\documentclass{article} %
\usepackage{iclr2027_conference,times}

\usepackage{amsmath,amsfonts,bm}

\def\eqref#1{equation~\ref{#1}}

\def\1{\bm{1}}

\DeclareMathAlphabet{\mathsfit}{\encodingdefault}{\sfdefault}{m}{sl}
\SetMathAlphabet{\mathsfit}{bold}{\encodingdefault}{\sfdefault}{bx}{n}

\usepackage{hyperref}
\usepackage{url}
\hypersetup{
  hidelinks,
  pdfkeywords={Large language models; SVD compression; Low-rank fine-tuning; Residual completion; Compression across budgets}
}

\usepackage{booktabs} 
\usepackage{tabularx}
\usepackage{graphicx}
\usepackage{enumitem}
\usepackage{multirow} 
\usepackage{placeins}
\usepackage[percent]{overpic}


\title{Learning to Be Truncated: Preparing LLM Checkpoints for SVD Compression}

\newcommand{\authoremail}[1]{{%
\hypersetup{pdfborder={0 0 1},urlbordercolor={0 1 1}}%
\texttt{\href{mailto:#1}{#1}}%
}}

\author{%
\hspace*{-\tabcolsep}%
\begin{tabularx}{\textwidth}{>{\centering\arraybackslash\normalfont}X>{\centering\arraybackslash\normalfont}X>{\centering\arraybackslash\normalfont}X}
\textbf{Chenyue Fang} & \textbf{Danila Mishin} & \textbf{Babak Falsafi}\\
\small{EcoCloud, EPFL} & \small{EcoCloud, EPFL} & \small{EcoCloud, EPFL}\\
\small{\authoremail{chenyue.fang@epfl.ch}} & \small{\authoremail{danila.mishin@epfl.ch}} & \small{\authoremail{babak.falsafi@epfl.ch}}\\
\end{tabularx}%
\hspace*{-\tabcolsep}%
}

\usepackage{xcolor}

\iclrfinalcopy
\begin{document}

\maketitle
\fancyhead{}
\renewcommand{\headrulewidth}{0pt}
\suppressfloats[t]

\begin{abstract}

Compression based on singular value decomposition (SVD) often requires
subsequent fine-tuning or weight correction to mitigate performance degradation.
Budget-specific optimization must be repeated for different compression ratios.
We introduce LLift, which prepares a pretrained model for SVD compression
across ratios without further fine-tuning.
LLift fine-tunes a low-rank approximation at a single ratio using the
original model as a teacher. We merge the learned updates into the original
weights while preserving the complementary components excluded from the
training forward pass. We apply SVD to the prepared model's target matrices
and reuse the decompositions at different truncation ranks.
On LLaMA-7B, we demonstrate accuracy gains across three compression ratios
using a single prepared model. The gains extend across model families,
model sizes, and SVD compression operators. We analyze gradients and an
early training update to provide local evidence that updates learned at
one ratio can benefit compression at other ratios.
These results show that LLift can improve compression across deployment
budgets without fine-tuning each compressed model separately.

\end{abstract}

\section{Introduction}
\label{sec:intro}

Deploying large language models across hardware with different memory,
latency, and compute budgets motivates compression methods that
produce models of different sizes from a single pretrained checkpoint.
Singular value decomposition (SVD) supports post hoc compression by replacing
weight matrices with low-rank factors of adjustable rank.
These models use standard dense matrix operations \citep{wang2025svdllm}.
However, truncation can degrade the model's predictions.
Existing methods reduce this degradation
through activation-aware reconstruction, rank allocation, and error correction
\citep{yuan2023asvd,wang2025dobisvddifferentiablesvdllm,wang2025svdllm,
hu2026saessvdselfadaptivesuppressionaccumulated,abbasi2026zerosumsvdbalancing}.

\begin{figure*}[t]
    \centering
    \includegraphics[width=\linewidth]{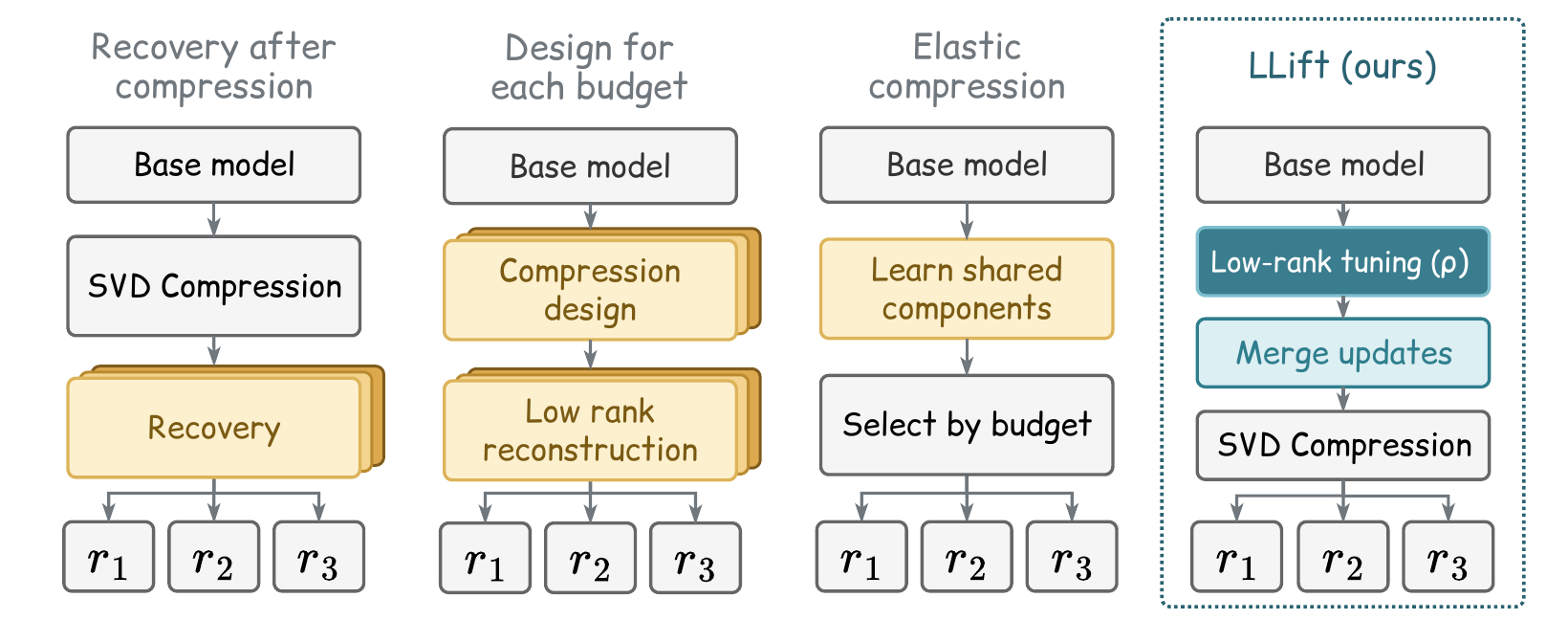}
    \caption{Compression workflows across budgets.
    Post-compression fine-tuning and budget-specific reconstruction repeat
    optimization for each budget, while elastic methods select from learned
    shared components.
    LLift fine-tunes a low-rank approximation at one preparation budget $\rho$
    and merges the learned updates into the original weights to prepare one model
    for compression across budgets.
    Stacked boxes indicate budget-specific work;
    $r_1,r_2,r_3$ denote deployment budgets.}
    \label{fig:teaser}
\end{figure*}

Figure~\ref{fig:teaser} compares different methods for producing compressed
models at multiple compression ratios.
Post-compression fine-tuning and budget-specific reconstruction improve
individual compressed models, but require repeating
optimization for each target. Elastic methods such as FlexRank and ACIP
amortize training across budgets using shared low-rank factors or a learned
component ordering \citep{zaccone2026flexrank,genzel2025choosemodelsizecompression}.
These methods instantiate deployment models from the learned representations.
We study a complementary goal: updating the original model, then truncating
shared SVDs of the prepared model's target matrices at different ranks.

Training with future compression in mind provides a foundation for this
goal. Compression-aware training encourages low-rank matrices
\citep{alvarez2017compressionaware}, while Decomposable-Net jointly trains
full- and low-rank networks \citep{yaguchi2021decomposable}. More recently,
Prehab prepares the original model by optimizing a task loss together with
spectral regularization \citep{qin2025lowrankprehabpreparingneural},
encouraging weight spectra that are more favorable to truncation.
Our approach instead uses prediction errors exposed by a low-rank model
to guide updates to the original model.

We introduce LLift, which fine-tunes a low-rank approximation of the original
model at one preparation budget. This approximation is the low-rank view:
its forward pass uses only the retained components, while the original
model provides supervision. After fine-tuning, we merge the learned update
into the original weights and preserve the components excluded from the
training forward pass. With fixed activation transforms, we apply SVD once
per target matrix of the prepared model and select different truncation ranks
from the shared decompositions. This separates the budget used to expose
compression error during training from the budget selected for deployment.

The challenge is to make updates learned at one budget useful when we
compress the prepared model at another. Exporting only the fine-tuned
student would discard components absent from the training view, including
information that larger deployment budgets could retain. LLift preserves
these complementary components while incorporating the learned correction
into the original weights. Recompressing the prepared model can then draw on
information outside the preparation subspace.

We show that preparing each model at one budget improves compression
quality at other evaluated budgets, with gains across model families,
sizes, and SVD operators. We analyze gradients at initialization and
identify shared descent directions as local evidence of cross-budget
compatibility.

Our contributions are threefold:
\begin{itemize}[leftmargin=*,itemsep=2pt,parsep=0pt,topsep=4pt,partopsep=0pt]
\item We introduce LLift, a model-preparation method that uses supervision
at one compression budget. We truncate shared SVDs of the prepared model's
target matrices to obtain models with competitive performance across
deployment budgets without further fine-tuning.
\item We show that updates learned by the low-rank student remain useful
after merging the updates and applying SVD to the prepared model.
Preserving the original complementary weights improves recompressed accuracy
over exporting the fine-tuned student alone at deployment.
\item We evaluate model preparation across model families, sizes, budgets,
and SVD compression operators, and analyze local cross-budget compatibility
through gradient alignment at initialization and recompression after the
first nonzero AdamW update to the original model.
\end{itemize}

\section{Related Work}

\paragraph{SVD-based LLM compression.}
Post hoc SVD compression has evolved from weight-space decomposition toward
activation-aware reconstruction, rank selection, and error correction.
Existing methods use activation-derived scaling~\citep{yuan2023asvd},
input-space whitening~\citep{wang2025svdllm}, differentiable rank
optimization~\citep{wang2025dobisvddifferentiablesvdllm,sundrani2025lowrankcompressionlanguagemodels},
cumulative-error suppression~\citep{hu2026saessvdselfadaptivesuppressionaccumulated},
loss-aware singular-component selection~\citep{abbasi2026zerosumsvdbalancing},
\leavevmode{activation-aware decomposition with dynamic rank allocation~\citep{qi2026swiftsvd},
Fisher-weighted decomposition with Kronecker-factored
approximations~\citep{chekalina2026generalizedfisherweightedsvdscalable},
and token-wise Fisher statistics with rank allocation using a calibrated
error model and integer linear programming~\citep{thoma2026lems}.}
Low-Rank Prehab encourages a more compressible weight spectrum before
applying SVD~\citep{qin2025lowrankprehabpreparingneural}, while NuMuon
promotes low-rank structure during pretraining through constrained optimizer
updates~\citep{dolatabadi2026numuonnuclearnormconstrainedmuoncompressible}.
PiSSA~\citep{meng2024pissa} adapts principal singular components for
parameter-efficient fine-tuning, retaining a frozen residual in its forward
pass.
LLift prepares a pretrained model by supervising its end-to-end
predictions through an activation-aware low-rank view. Its objective targets
compressed-model behavior without explicitly regularizing the weight spectrum,
and we evaluate the prepared model with SVD-LLM, ZS-SVD, and
KFAC-SVD with LEMS rank allocation at deployment.

\paragraph{Multi-budget and elastic low-rank deployment.}
Nested and elastic models support multiple deployment budgets through shared
representations or subnetworks, with smaller configurations contained in
larger ones~\citep{kusupati2022matryoshka,devvrit2024matformernestedtransformerelastic,
cai2024flextronmanyinoneflexiblelarge,taghibakhshi2025nemotronelasticefficientmanyinone}.
For low-rank deployment, selecting subsets of shared components adjusts
model size. FlexRank combines shared low-rank factors with
budget-specific rank configurations and trains the resulting nested models
across budgets~\citep{zaccone2026flexrank}. ACIP learns a global importance
ordering over SVD components together with a shared low-rank correction,
allowing deployment at different model sizes by truncating the
ordering~\citep{genzel2025choosemodelsizecompression}. Both methods organize
multi-budget deployment around a shared low-rank representation or component
ordering and use these representations to instantiate deployment models
directly at different budgets.

LLift instead prepares the original model through supervision
of a low-rank view at a single budget and merges the learned updates into
the original weights. At deployment, we obtain different model sizes by
selecting truncation ranks from shared SVDs of the prepared model's target
matrices. We test whether this preparation improves performance at other
compression budgets.

\section{Preliminaries}
\label{sec:prelim}

\paragraph{Activation-aware low-rank approximation.}
For a linear layer with weight
$W\in\mathbb R^{d_{\mathrm{out}}\times d_{\mathrm{in}}}$
and calibration activations $X\in\mathbb R^{d_{\mathrm{in}}\times N}$,
activation-aware compression seeks a rank-$k$ weight approximation
that best preserves the original layer outputs on these calibration activations:
\begin{equation}
\min_{\operatorname{rank}(\widehat W)\leq k}
\left\|(W-\widehat W)X\right\|_F^2.
\label{eq:activation_objective}
\end{equation}
Following the activation-whitening formulation of SVD-LLM
\citep{wang2025svdllm}, we use $XX^\top$ to summarize the calibration
activations' second-order statistics. Adding $\epsilon I$, with
$\epsilon>0$, makes this matrix positive definite. We can therefore obtain
an invertible factor $T$ satisfying $TT^\top=XX^\top+\epsilon I$.
We can compute $T$ by Cholesky decomposition and obtain the following identity:
\begin{equation}
\left\|(W-\widehat W)T\right\|_F^2
=\left\|(W-\widehat W)X\right\|_F^2
+\epsilon\left\|W-\widehat W\right\|_F^2.
\label{eq:weighted_objective}
\end{equation}
Approximating $WT$ minimizes the output reconstruction error
together with a weight reconstruction penalty. We decompose
$WT=U\Sigma V^\top$, with singular values in descending order.
Since $T^{-1}$ preserves rank, we solve the damped objective by truncating
$WT$ and multiplying by $T^{-1}$:
\begin{equation}
\widehat W^{(k)}=U_{:k}\Sigma_{:k}V_{:k}^\top T^{-1},
\label{eq:weighted_svd}
\end{equation}
Here $U_{:k}$ and $V_{:k}$ contain the first $k$ columns of $U$ and $V$,
respectively. We retain the leading $k\times k$ diagonal block of $\Sigma$
in $\Sigma_{:k}$, corresponding to the $k$ largest singular values of $WT$.

\paragraph{Nested rank truncation.}
For a fixed checkpoint and activation transform, the components retained at
rank $k_1$ are a subset of those retained at rank $k_2$ whenever $k_1<k_2$.
A single SVD can therefore produce multiple model sizes by retaining
different numbers of leading components. This nesting permits reuse of the
decomposition but does not by itself guarantee task performance at unseen
budgets. LLift uses activation-aware SVD to initialize its low-rank training
view. At deployment, we apply SVD to the prepared model's transformed target
matrices and select different truncation ranks from these shared
decompositions to obtain models at each deployment budget.

\section{Method}
\label{sec:method}

\begin{figure*}[t]
    \centering
    \includegraphics[width=\linewidth]{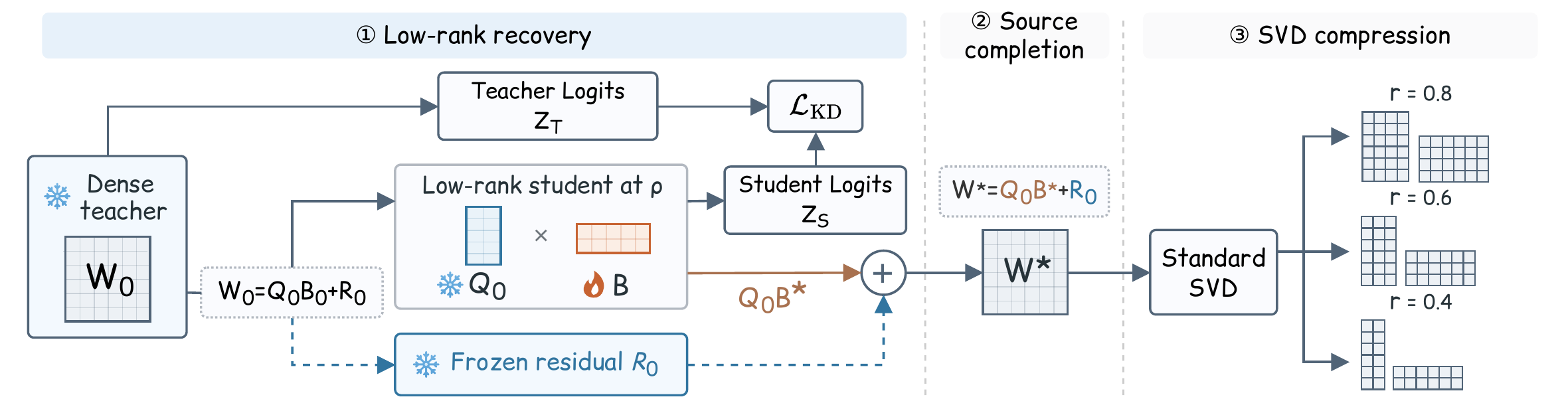}
    \caption{Fine-tuning a low-rank view and merging its updates into the original model.
    LLift uses the frozen original model to supervise a student whose target layer
    uses $Q_0B$ at preparation anchor $\rho$, with $Q_0$ fixed and $B$ trainable.
    Snowflakes and flames denote frozen and trainable quantities.
    After training, we merge the update $Q_0(B^\star-B_0)$ into $W_0$,
    yielding $W^\star=Q_0B^\star+R_0$, where $R_0=W_0-Q_0B_0$ is the frozen
    complementary component that bypasses the training forward pass.
    With a fixed activation transform $T$, we decompose $W^\star T$ once
    and truncate the shared decomposition to the ranks specified by $r$.
    The diagram shows one target matrix; the prepared model also retains
    the embeddings, output head, and normalization parameters adapted
    during student fine-tuning.}
    \label{fig:overview}
\end{figure*}

LLift prepares a model for SVD compression across deployment budgets.
We fine-tune a low-rank student at one preparation anchor, using fixed
subspaces derived from the original weights and activation statistics. We merge
these updates into the original weights, preserving the complementary
components excluded from the student forward pass. For deployment, we
truncate shared SVDs of the prepared model's transformed target matrices
at the ranks specified by each budget (Figure~\ref{fig:overview}).

\subsection{Learning through a Low-Rank View}
\label{sec:method_setup}
\label{sec:method_LoLift}

Activation-aware SVD constructs each compressed layer from local weight and
activation statistics, while the quality of the resulting model depends on
the joint effect of compression across layers. LLift therefore exposes all
target layers to their low-rank views simultaneously and uses the pretrained
model to supervise the resulting end-to-end behavior. We use one preparation
anchor to define the training view and select deployment ranks when
compressing the prepared model.

\begin{samepage}
Let $f(x;\Theta_0)$ denote the original model's logits for input $x$, with target matrices
$\{W_{\ell,0}\}_{\ell\in\mathcal S}$ and remaining parameters $\Phi_0$.
We use $\rho$ for preparation and $r$ for deployment;
both specify nominal retained-parameter fractions in the target matrices.
We assign $\rho$ uniformly across target matrices.
A rank-$k_\ell$ factorization stores
$k_\ell(d_{\mathrm{out},\ell}+d_{\mathrm{in},\ell})$ parameters instead of
$d_{\mathrm{out},\ell}d_{\mathrm{in},\ell}$, so we choose
\begin{equation}
k_\ell(\rho)=\max\!\left\{1,
\left\lfloor
\frac{\rho\,d_{\mathrm{out},\ell}d_{\mathrm{in},\ell}}
{d_{\mathrm{out},\ell}+d_{\mathrm{in},\ell}}
\right\rfloor\right\}.
\label{eq:rank_budget}
\end{equation}
The rank $k_\ell(\rho)$ determines the dimensions of the two factors used
in the training forward pass. Uniform parameter retention, subject to rank
rounding, does not imply the same retained fraction of singular directions
across target matrices with different input and output dimensions.
\end{samepage}

The initial view preserves the leading activation-aware components of the
pretrained weights. We estimate $T_\ell^{\mathrm{train}}$ from the original
model on calibration text and decompose
$W_{\ell,0}T_\ell^{\mathrm{train}}=U_\ell\Sigma_\ell V_\ell^\top$ as in
Eq.~\ref{eq:weighted_svd}. We use the retained left singular vectors as a
fixed basis and project the weights:
\begin{equation}
Q_{\ell,0}=U_{\ell,:k_\ell(\rho)},
\qquad
B_{\ell,0}=Q_{\ell,0}^\top W_{\ell,0}.
\label{eq:factor_initialization}
\end{equation}
Here $Q_{\ell,0}\in\mathbb R^{d_{\mathrm{out},\ell}\times k_\ell(\rho)}$
and $B_{\ell,0}\in\mathbb R^{k_\ell(\rho)\times d_{\mathrm{in},\ell}}$.
The product $Q_{\ell,0}B_{\ell,0}=Q_{\ell,0}Q_{\ell,0}^\top W_{\ell,0}$
is the rank-$k_\ell(\rho)$ projection induced by the activation-aware
decomposition in Eq.~\ref{eq:weighted_svd}. We associate each coordinate
matrix $B_\ell$ with a corresponding model weight parameterized as
\[
W_\ell(B_\ell)=W_{\ell,0}
+Q_{\ell,0}(B_\ell-B_{\ell,0}).
\]
Since $Q_{\ell,0}^\top Q_{\ell,0}=I$, its projection onto the fixed column
space of $Q_{\ell,0}$ satisfies
$Q_{\ell,0}Q_{\ell,0}^\top W_\ell(B_\ell)=Q_{\ell,0}B_\ell$.
The student evaluates only the projected map
$h\mapsto Q_{\ell,0}(B_\ell h)$. The complementary component of the
original weight remains unchanged and does not participate in
this forward pass. Preparation therefore does not require optimizing
full target matrices. We keep $Q_{\ell,0}$ fixed.

The original model serves as a frozen teacher. Let
$z_t^{\mathrm T}=f(x;\Theta_0)_t$ denote its logits at position $t$ and
$z_t^{\mathrm S}(\mathcal B,\Phi)$ those of the student with
target weights $\{Q_{\ell,0}B_\ell\}_{\ell\in\mathcal S}$, where
$\mathcal B=\{B_\ell\}_{\ell\in\mathcal S}$ and $\Phi$ collects other
student parameters. We minimize the token-level student-to-teacher KL loss:
\begin{equation}
\mathcal L_{\mathrm{LLift}}(\mathcal B,\Phi)=
\mathbb E_{x\sim\mathcal D_{\mathrm{train}}}
\left[
\frac{1}{|\Omega(x)|}\sum_{t\in\Omega(x)}
\mathrm{KL}\!\left(
p_\tau(z_t^{\mathrm S}(\mathcal B,\Phi))\,\middle\|\,
p_\tau(z_t^{\mathrm T})
\right)
\right],
\label{eq:LoLift_loss}
\end{equation}
where $p_\tau(z)=\operatorname{softmax}(z/\tau)$ and $\Omega(x)$ excludes
padded positions. Supervision at the final logits accounts for the combined
effect of compressing all target layers. \leavevmode{During training, we omit the original
complementary components $W_{\ell,0}-Q_{\ell,0}B_{\ell,0}$ from the student
forward pass. This exposes prediction errors caused by compression and provides
a learning signal for adapting the retained coordinates.} We initialize
$(\mathcal B,\Phi)\leftarrow(\mathcal B_0,\Phi_0)$ and jointly optimize
these parameters, including the embeddings, output head, and normalization
parameters. We keep the teacher and bases $Q_{\ell,0}$ fixed.

\subsection{Constructing the Prepared Model}
\label{sec:method_residual}

We update the original weights by merging the coordinate change
$B_\ell^\star-B_{\ell,0}$ into $W_{\ell,0}$ through $Q_{\ell,0}$,
preserving the original components that the training forward pass excludes.
We can equivalently combine the fine-tuned low-rank factors with the
original truncation residual.

\begin{samepage}
We retain the original truncation residual
\begin{equation}
R_{\ell,0}=W_{\ell,0}-Q_{\ell,0}B_{\ell,0},
\qquad Q_{\ell,0}^\top R_{\ell,0}=0.
\label{eq:initial_residual}
\end{equation}
We write each target weight in the prepared model in two equivalent forms:
\begin{equation}
\begin{aligned}
W_\ell^\star
&=W_{\ell,0}+Q_{\ell,0}(B_\ell^\star-B_{\ell,0})\\
&=Q_{\ell,0}B_\ell^\star+R_{\ell,0}.
\end{aligned}
\label{eq:residual_restoration}
\end{equation}
\end{samepage}
Reattaching $R_{\ell,0}$ yields the minimum-change completion of the
fine-tuned coordinates:
\begin{equation}
W_\ell^\star
=\operatorname*{arg\,min}_{W:\,Q_{\ell,0}^\top W=B_\ell^\star}
\|W-W_{\ell,0}\|_F^2.
\label{eq:minimum_change_completion}
\end{equation}
The constraint fixes the trained component, and the nearest completion
preserves the original complement. We prove this minimum-distance property
in Appendix~\ref{app:minimum_change}; it does not establish task-loss
optimality. With no adaptation,
Eq.~\ref{eq:residual_restoration} recovers $W_{\ell,0}$.
After adaptation, projection onto the fixed basis still recovers the trained view:
$Q_{\ell,0}Q_{\ell,0}^\top W_\ell^\star=Q_{\ell,0}B_\ell^\star$.

The fixed $R_{\ell,0}$ requires no gradients or optimizer states.
We reattach it only when exporting the prepared model. The target
matrices require optimizer states only for $B_\ell$. The exported model
is $\Theta^\star=(\{W_\ell^\star\}_{\ell\in\mathcal S},\Phi^\star)$,
including the adapted remaining parameters.
In Appendix~\ref{app:projection_derivation}, we derive when factorized and
full-matrix parameterizations induce identical updates to the target weights.

\subsection{Recompression beyond the Training Budget}
\label{sec:method_deploy}

The intended benefit of preparation is improved quality after recompressing
the prepared model, including at budgets not seen during preparation. Deployment
compression acts on the completed weights
$W_\ell^\star=Q_{\ell,0}B_\ell^\star+R_{\ell,0}$, so both the learned
update and the preserved complementary component can influence the
singular directions retained by SVD at each deployment budget.

The orthogonality condition $Q_{\ell,0}^\top R_{\ell,0}=0$ refers to the
fixed preparation basis. For a deployment activation transform $T$, it
does not imply $B_\ell^\star TT^\top R_{\ell,0}^\top=0$.
The cross terms in $(W_\ell^\star T)(W_\ell^\star T)^\top$ can therefore
couple the fine-tuned component to the preserved complement. Applying SVD
to $W_\ell^\star T$ can select left singular directions that differ from
the preparation basis, allowing the complement to affect compression
even when $r\leq\rho$. At $r=\rho$, the recompressed model consequently
need not reproduce the predictions of the low-rank student trained at that
same preparation budget.

\leavevmode{Restoring $R_{\ell,0}$ allows deployment SVD to retain components absent
from $Q_{\ell,0}B_\ell^\star$. We show that this can reduce reconstruction
error relative to the original teacher weight $W_{\ell,0}$ after compression
at the same rank, and derive sufficient conditions for improvement at any
deployment rank. Above the preparation rank, the restored residual supplies
additional directions that the trained low-rank weight alone cannot provide.
Appendix~\ref{app:residual_reconstruction} gives the derivations.}

We hold the deployment activation transforms $T_\ell^{\mathrm{deploy}}$
fixed across budgets and apply SVD once to each
$W_\ell^\star T_\ell^{\mathrm{deploy}}$. For budget $r$, we retain the
first $k_\ell(r)$ components of these shared decompositions and construct
the layer factors as in Eq.~\ref{eq:weighted_svd}.
Let $\mathcal C_r(\Theta;\mathcal D_{\mathrm{cal}}^{\mathrm{deploy}})$
denote compression with these fixed transforms. Applying this operator to
the prepared model yields the deployment models
\begin{equation}
\Theta_r=\mathcal C_r\!\left(
\Theta^\star;\mathcal D_{\mathrm{cal}}^{\mathrm{deploy}}
\right),\qquad r\in\mathcal R.
\label{eq:deployment}
\end{equation}
Each budget requires its own truncation ranks and factor materialization,
while reusing the SVD of each target matrix without additional fine-tuning.
Section~\ref{sec:source_comparison} examines cross-budget transfer through
the alignment of the ratio-specific fine-tuning gradients at initialization.

\section{Experiments}
\label{sec:experiments}

We test whether preparation at one anchor improves compression across
budgets, examine residual restoration, and assess generality
across SVD operators, model families, and scales.

\subsection{Experimental Setup}
\label{sec:experimental_setup}

\paragraph{Preparation.}
LLaMA-7B is our reference model.
Unless otherwise stated, we prepare each model once at $\rho=0.6$
on the WikiText-2 (WT2)~\citep{wikitext} training split using KL distillation from the original
model. We construct the fixed preparation bases from original-model
activations on WT2.
We use student-to-teacher KL as the default objective and report CE and
KL+CE controls in Appendix~\ref{app:objective_anchor}. In particular, removing
KL supervision and training with CE alone substantially increases
post-compression perplexity, especially at smaller deployment budgets.

\paragraph{Deployment.}
We use SVD-LLM to obtain deployment models at
$r \in \{0.8, 0.6, 0.4\}$ without further fine-tuning.
For matched comparisons, we estimate deployment activation transforms
from the original model on a separate WT2/Alpaca~\citep{alpaca} calibration set
and hold them fixed across models and budgets. We reuse each model's SVDs across
all three budgets by varying the truncation ranks.
Re-estimating these transforms from the prepared model changes accuracy
only modestly and provides no consistent perplexity benefit
(Appendix~\ref{app:deployment_recalibration}).
Following Eq.~\ref{eq:rank_budget}, $r$ denotes the nominal ratio of factor
parameters to original target-matrix parameters.

\paragraph{Evaluation.}
We report held-out perplexity on WT2, Penn Treebank (PTB)~\citep{marcus-etal-1993-building},
and C4~\citep{2020t5}, together with the equally weighted mean of plain
zero-shot accuracy (Avg.) on seven tasks:
OpenBookQA, ARC-Easy, ARC-Challenge, WinoGrande, HellaSwag, PIQA,
and MathQA
\citep{OpenBookQA2018,arc,sakaguchi2019winogrande,zellers2019hellaswag,
bisk2019piqareasoningphysicalcommonsense,amini-etal-2019-mathqa}.
\subsection{Quality Across Deployment Budgets}
\label{sec:quality_cost}
\label{sec:quality_across_budgets}

\begin{table}[t]
\centering
\scriptsize
\setlength{\tabcolsep}{2.6pt}
\caption{SVD compression on LLaMA-7B. Baselines use official calibration settings. WT2, PTB, and C4 report perplexity ($\downarrow$). Avg. is seven-task mean plain accuracy ($\uparrow$). Bold and underlining mark the best and second-best in these columns per budget. $\dagger$: budget-specific optimization.
\leavevmode{LLift uses one WT2 preparation run shared across deployment budgets.
Appendix~\ref{app:matched_recovery} compares budget-specific recovery under
matched training conditions.}}

\label{tab:main_llama7b_full}

\resizebox{\linewidth}{!}{%
\begin{tabular}{llccc|cccccccc}
\toprule
$r$ & Method & WT2$\downarrow$ & PTB$\downarrow$ & C4$\downarrow$ &
OBQA$\uparrow$ & ARC-e$\uparrow$ & ARC-c$\uparrow$ & WinoG$\uparrow$ &
HellaS$\uparrow$ & PIQA$\uparrow$ & MathQA$\uparrow$ & Avg.$\uparrow$ \\
\midrule
\multicolumn{2}{l}{Uncompressed} & 5.68 & \leavevmode{8.35} & 7.34 & 0.31 & 0.75 & 0.42 & 0.69 & 0.56 & 0.78 & 0.26 & 0.54 \\
\midrule
0.8 & ASVD & 11.14 & \leavevmode{16.55} & 15.93 & 0.25 & 0.53 & 0.27 & 0.64 & 0.41 & 0.68 & 0.24 & 0.43 \\
 & SVD-LLM & 7.94 & \leavevmode{16.22} & 15.84 & 0.22 & 0.58 & 0.29 & 0.63 & 0.43 & 0.69 & 0.24 & 0.44 \\
 & Dobi-SVD$^\dagger$ & 8.54 & \leavevmode{14.83} & \underline{10.01} & 0.26 & 0.59 & 0.31 & 0.66 & 0.44 & 0.70 & 0.23 & 0.46 \\
 & SAES-SVD$^\dagger$ & 7.17 & \leavevmode{15.16} & 13.77 & 0.29 & 0.68 & 0.36 & 0.65 & 0.45 & 0.75 & 0.25 & 0.49 \\
 & ZS-SVD & 6.74 & \leavevmode{11.87} & 10.74 & 0.31 & 0.70 & 0.37 & 0.68 & 0.49 & 0.74 & 0.24 & \underline{0.50} \\
 & ZS-SVD 5$\times^\dagger$ & \underline{6.43} & \leavevmode{\underline{11.17}} & 10.42 & 0.32 & 0.72 & 0.39 & 0.67 & 0.50 & 0.75 & 0.25 & \textbf{0.51} \\
 & Swift-SVD$^*$ & \leavevmode{11.64} & \leavevmode{14.74} & \leavevmode{11.76} & \leavevmode{0.24} & \leavevmode{0.63} & \leavevmode{0.31} & \leavevmode{0.64} & \leavevmode{0.44} & \leavevmode{0.71} & \leavevmode{0.23} & \leavevmode{0.46} \\
 & LEMS & \leavevmode{\textbf{6.23}} & \leavevmode{\textbf{10.27}} & \leavevmode{\textbf{9.15}} & \leavevmode{0.29} & \leavevmode{0.73} & \leavevmode{0.38} & \leavevmode{0.67} & \leavevmode{0.52} & \leavevmode{0.76} & \leavevmode{0.24} & \leavevmode{\textbf{0.51}} \\
 & \textbf{LLift} & 7.49 & \leavevmode{11.72} & 11.15 & 0.30 & 0.67 & 0.37 & 0.66 & 0.52 & 0.74 & 0.25 & \underline{0.50} \\
\midrule
0.6 & ASVD & 1407 & \leavevmode{3292} & 1109 & 0.13 & 0.28 & 0.22 & 0.48 & 0.26 & 0.55 & 0.19 & 0.30 \\
 & SVD-LLM & 13.11 & \leavevmode{63.75} & 49.83 & 0.19 & 0.42 & 0.25 & 0.58 & 0.33 & 0.60 & 0.21 & 0.37 \\
 & Dobi-SVD$^\dagger$ & 13.54 & \leavevmode{46.38} & 23.54 & 0.22 & 0.41 & 0.27 & 0.58 & 0.34 & 0.61 & 0.23 & 0.38 \\
 & SAES-SVD$^\dagger$ & 10.42 & \leavevmode{45.13} & 32.79 & 0.23 & 0.50 & 0.29 & 0.62 & 0.36 & 0.65 & 0.23 & 0.41 \\
 & ZS-SVD & 11.44 & \leavevmode{43.19} & 34.13 & 0.23 & 0.52 & 0.26 & 0.62 & 0.35 & 0.64 & 0.22 & 0.41 \\
 & ZS-SVD 5$\times^\dagger$ & 9.45 & \leavevmode{36.52} & 26.20 & 0.23 & 0.56 & 0.29 & 0.62 & 0.38 & 0.66 & 0.22 & 0.42 \\
 & Swift-SVD$^*$ & \leavevmode{36.87} & \leavevmode{56.15} & \leavevmode{25.20} & \leavevmode{0.16} & \leavevmode{0.40} & \leavevmode{0.22} & \leavevmode{0.54} & \leavevmode{0.30} & \leavevmode{0.61} & \leavevmode{0.22} & \leavevmode{0.35} \\
 & LEMS & \leavevmode{\underline{9.22}} & \leavevmode{\underline{21.46}} & \leavevmode{\underline{19.29}} & \leavevmode{0.22} & \leavevmode{0.60} & \leavevmode{0.29} & \leavevmode{0.61} & \leavevmode{0.39} & \leavevmode{0.67} & \leavevmode{0.22} & \leavevmode{\underline{0.43}} \\
 & \textbf{LLift} & \textbf{8.40} & \leavevmode{\textbf{15.96}} & \textbf{14.77} & 0.25 & 0.63 & 0.31 & 0.62 & 0.45 & 0.71 & 0.25 & \textbf{0.46} \\
\midrule
0.4 & ASVD & 57057 & \leavevmode{45218} & 43036 & 0.12 & 0.26 & 0.21 & 0.49 & 0.26 & 0.53 & 0.18 & 0.29 \\
 & SVD-LLM & 53.74 & \leavevmode{438.58} & 345.49 & 0.14 & 0.28 & 0.22 & 0.50 & 0.27 & 0.55 & 0.21 & 0.31 \\
 & Dobi-SVD$^\dagger$ & 46.18 & \leavevmode{238.91} & 190.62 & 0.15 & 0.31 & 0.20 & 0.52 & 0.28 & 0.54 & 0.22 & 0.32 \\
 & SAES-SVD$^\dagger$ & 22.01 & \leavevmode{116.83} & 93.97 & 0.16 & 0.33 & 0.25 & 0.52 & 0.30 & 0.54 & 0.23 & \underline{0.34} \\
 & ZS-SVD & 45.17 & \leavevmode{334.85} & 212.57 & 0.15 & 0.31 & 0.21 & 0.54 & 0.28 & 0.55 & 0.21 & 0.32 \\
 & ZS-SVD 5$\times^\dagger$ & \underline{20.41} & \leavevmode{163.58} & 88.56 & 0.17 & 0.37 & 0.22 & 0.55 & 0.30 & 0.57 & 0.21 & \underline{0.34} \\
 & Swift-SVD$^*$ & \leavevmode{257.70} & \leavevmode{400.56} & \leavevmode{143.85} & \leavevmode{0.13} & \leavevmode{0.27} & \leavevmode{0.21} & \leavevmode{0.49} & \leavevmode{0.26} & \leavevmode{0.55} & \leavevmode{0.21} & \leavevmode{0.30} \\
 & LEMS & \leavevmode{21.93} & \leavevmode{\underline{104.05}} & \leavevmode{\underline{83.94}} & \leavevmode{0.16} & \leavevmode{0.37} & \leavevmode{0.22} & \leavevmode{0.55} & \leavevmode{0.30} & \leavevmode{0.59} & \leavevmode{0.22} & \leavevmode{\underline{0.34}} \\
 & \textbf{LLift} & \textbf{13.48} & \leavevmode{\textbf{53.47}} & \textbf{32.24} & 0.17 & 0.48 & 0.24 & 0.57 & 0.34 & 0.62 & 0.22 & \textbf{0.38} \\
\bottomrule
\end{tabular}%
}
\end{table}

\paragraph{Comparison with SVD compression methods.}
Table~\ref{tab:main_llama7b_full} compares LLift with
ASVD~\citep{yuan2023asvd}, SVD-LLM~\citep{wang2025svdllm},
Dobi-SVD~\citep{wang2025dobisvddifferentiablesvdllm},
SAES-SVD~\citep{hu2026saessvdselfadaptivesuppressionaccumulated},
ZS-SVD~\citep{abbasi2026zerosumsvdbalancing},
Swift-SVD$^*$~\citep{qi2026swiftsvd}, and
LEMS~\citep{thoma2026lems}, which denotes KFAC-SVD with LEMS rank allocation.
LLift raises mean accuracy from the reported SVD-LLM scores of
$0.44/0.37/0.31$ to $0.50/0.46/0.38$ at $r=0.8/0.6/0.4$.
At $r=0.6/0.4$, it achieves the highest mean accuracy and lowest
perplexity on all three corpora among the methods in the table.
LEMS achieves $0.51/0.43/0.34$ accuracy at
$r=0.8/0.6/0.4$. It outperforms LLift at $r=0.8$ on mean
accuracy and all three perplexities, while LLift performs better
on these metrics at $r=0.6/0.4$.

\begin{table}[t]
\centering
\begin{minipage}[t]{0.485\textwidth}
    \vspace{0pt}
    \centering\small
\caption{Elastic methods. Seven-task average accuracy ($\uparrow$). $\Delta$ is a rank-32 correction whose parameters are excluded from $r$.}
\label{tab:elastic}
\footnotesize\setlength{\tabcolsep}{6pt}
\renewcommand{\arraystretch}{1}
\begin{tabular}{@{}p{0.30\linewidth}ccc@{}}
\toprule
Method & $r=0.8$ & $r=0.6$ & $r=0.4$ \\
\midrule
ACIP & 0.49 & 0.40 & 0.33 \\
ACIP ($+\Delta$) & 0.48 & 0.45 & 0.38 \\
FlexRank & 0.49 & 0.46 & 0.41 \\
\textbf{LLift} & 0.50 & 0.46 & 0.38 \\
\bottomrule
\end{tabular}

\end{minipage}
\hfill
\begin{minipage}[t]{0.485\textwidth}
    \vspace{0pt}
    \centering\small
\setlength{\tabcolsep}{3pt}
\caption{Model preparation and residual restoration under matched \texttt{mix75} calibration. Seven-task average accuracy ($\uparrow$).}
\label{tab:controlled_sources}
\footnotesize\renewcommand{\arraystretch}{1}
\begin{tabular*}{\linewidth}{@{\extracolsep{\fill}}lccc@{}}
\toprule
Model & $r=0.8$ & $r=0.6$ & $r=0.4$ \\
\midrule
Original & 0.44 & 0.40 & 0.31 \\
Spectral regularization & 0.48 & 0.41 & 0.33 \\
Fine-tuned $Q_0B^\star$ only & 0.44 & 0.44 & 0.37 \\
\textbf{LLift}: $Q_0B^\star+R_0$ & 0.50 & 0.46 & 0.38 \\
\bottomrule
\end{tabular*}

\end{minipage}
\end{table}

\paragraph{Comparison with elastic methods.}
Table~\ref{tab:elastic} compares LLift, FlexRank, and ACIP
with and without its rank-32 correction $\Delta$.
ACIP without $\Delta$ obtains $0.49/0.40/0.33$ accuracy at $r=0.8/0.6/0.4$,
versus $0.50/0.46/0.38$ for LLift.
FlexRank obtains accuracies of $0.49/0.46/0.41$ at these budgets,
matching LLift in mean accuracy at $r=0.6$ and outperforming it at $r=0.4$.
FlexRank's preparation takes $27.0$ A100 GPU-hours, mostly for per-layer rank
profiling, compared with $2.07$ for LLift (Appendix~\ref{app:preparation_resources}).

\paragraph{Budget-specific fine-tuning.}
\leavevmode{We compare LLift with budget-specific SVD-LLM
recovery using the same WT2 training data, KL objective,
optimizer settings, and deployment calibration. Each
recovery model receives the training exposure of one LLift
preparation run. On LLaMA-7B and Qwen3-8B, LLift achieves
broadly comparable seven-task mean accuracy at
$r=0.8/0.6$, while recovery yields lower WT2 and C4
perplexity at all three budgets. At $r=0.4$, recovery
also achieves higher mean accuracy: $0.40$ versus $0.38$
on LLaMA-7B and $0.39$ versus $0.34$ on Qwen3-8B
(Table~\ref{tab:matched_recovery}).
For the three deployment budgets, LLift incurs a total
pipeline cost of $3.13$ A100 GPU-hours on LLaMA-7B,
compared with $5.10$ for separate recovery
(Table~\ref{tab:matched_recovery_costs}).
These results highlight the trade-off:
LLift amortizes preparation across deployment budgets
without additional fine-tuning, whereas budget-specific
recovery can achieve higher per-model quality,
particularly under stronger compression, at the cost
of repeated optimization.}

\subsection{Model Preparation and Cross-Budget Transfer}
\label{sec:source_comparison}
\label{sec:completion_and_transfer}

\paragraph{Relation to Low-Rank Prehab.}
Low-Rank Prehab~\citep{qin2025lowrankprehabpreparingneural} prepares
a model before SVD, optimizing a task loss on the original model
with spectral rank regularization. LLift supervises a low-rank view at one
anchor and restores the original complement.
\leavevmode{We compare our Prehab reimplementation
with LLift at matched training exposure in
Appendix~\ref{app:prehab_relation}. However, the
reimplementation's uncompressed seven-task mean
accuracy already drops from $0.54$ to $0.35$.
We therefore treat this comparison as a reproduction diagnostic.}
A separate control using
truncated Schatten spectral regularization improves mean accuracy over
Original but remains below LLift at all three budgets
(Table~\ref{tab:controlled_sources}). We detail the control's objective
and training setup in Appendix~\ref{app:spectral_control}.

\paragraph{Comparison with Decomposable-Net.}
\leavevmode{We evaluate a LLaMA-7B adaptation of
Decomposable-Net~\citep{yaguchi2021decomposable} under matched WT2 training
exposure of approximately $11.5$M token positions.
With the same SVD-LLM deployment operator, Decomposable-Net achieves mean
accuracies of $0.47/0.42/0.34$ at retention ratios of $0.8/0.6/0.4$,
compared with $0.50/0.46/0.38$ for LLift.
Appendix~\ref{app:decomposable_net} provides the adaptation details and
results with both deployment operators.}

\paragraph{Model preparation and residual restoration.}
In Table~\ref{tab:controlled_sources}, LLift raises mean accuracy
from $0.44/0.40/0.31$ to $0.50/0.46/0.38$ at $r=0.8/0.6/0.4$.
A single prepared model therefore benefits deployment budgets both
above and below the preparation anchor.

To isolate residual restoration, we apply SVD-LLM to the fine-tuned student
$Q_0B^\star$ and the prepared model $Q_0B^\star+R_0$ from the same
preparation run, with identical adapted non-target parameters, deployment
transforms, and ranks. Restoring $R_0$ improves seven-task mean accuracy
from $0.44/0.44/0.37$ to $0.50/0.46/0.38$ at $r=0.8/0.6/0.4$.

\leavevmode{Our paired diagnostic shows that residual restoration reduces measured
teacher reconstruction error for all 224 target matrices at each deployment
budget. The sufficient condition in Appendix~\ref{app:residual_reconstruction}
certifies improvement for $100.0\%/73.7\%/79.9\%$ of these matrices at
$r=0.8/0.6/0.4$. The remaining matrices also improve empirically, indicating
that the sufficient condition is conservative.
Appendix~\ref{app:residual_diagnostic} provides the detailed diagnostic results.}

\paragraph{Role of non-target parameters.}
We freeze LLaMA-7B's embeddings, output head, and normalization parameters
and train only the factor coordinates.
Both variants' mean accuracies round to $0.50/0.46/0.38$,
whereas freezing these parameters increases WT2
perplexity by $12$--$13\%$ and C4 perplexity by approximately $4$--$5\%$
across the three deployment budgets (Appendix~\ref{app:dense_freeze}).
We find that adapting these parameters primarily improves
perplexity in this control.

\paragraph{Spectral energy retention.}
At the reported precision, the original and prepared models differ by at
most $2\times10^{-4}$ in retained whitened spectral energy fractions across
the three budgets (Appendix~\ref{app:lolift_mechanism}). This suggests that
retained energy alone does not fully explain compression quality.

\begin{figure}[t]
    \centering
    \includegraphics[width=\linewidth]{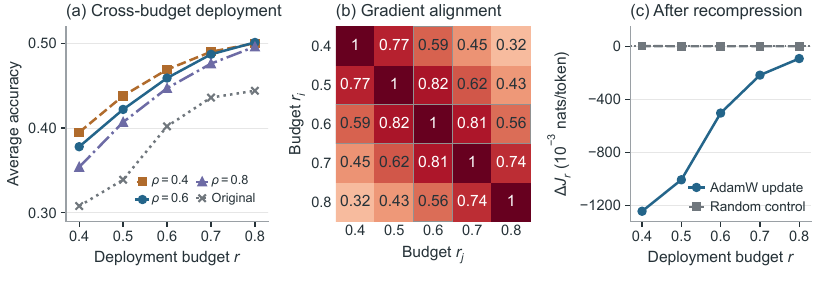}
\caption{
Cross-budget transfer and local compatibility on LLaMA-7B.
(a) Seven-task mean accuracy across deployment budgets for models prepared
at different anchors; we include the original model as a reference.
(b) Cosine similarities of fixed-projector teacher-KL gradients
at initialization.
(c) Change in held-out student-to-teacher KL after updating the original model
and recompressing it with SVD. We compare the first nonzero AdamW update
($s=1$) with a norm-matched random update. Negative values indicate lower
KL than when we compress the original model.
}
    \label{fig:lolift_mechanism}
\end{figure}

\paragraph{Preparation anchor.}
Preparing at $\rho=0.4$ improves accuracy at $r=0.6/0.4$
relative to the default configuration (Figure~\ref{fig:lolift_mechanism}(a)).
The default $\rho=0.6$ yields the lowest WT2 perplexity
at all three deployment budgets among the tested anchors
(Appendix~\ref{app:objective_anchor}), revealing a tradeoff between task accuracy
and language-modeling quality.
The anchor-matched points ($\rho=r$) in Figure~\ref{fig:lolift_mechanism}(a)
provide budget-specific preparation references. The single default
$\rho=0.6$ prepared model remains competitive, with accuracy differences of
$+0.50/0/-1.66$ percentage points at $r=0.8/0.6/0.4$.
Figure~\ref{fig:lolift_mechanism}(a) further evaluates intermediate deployment
budgets $r=0.5$ and $0.7$, which we do not use as preparation anchors,
providing denser evidence of transfer away from the preparation budget.

\paragraph{Local cross-budget compatibility.}
\label{sec:lolift_mechanism}
We probe teacher-KL objectives at the original checkpoint while holding
activation transforms and budget-specific projectors $P_r$ fixed.
Let $g_r=\nabla_W\widetilde{\mathcal L}_r(W_0;P_r)$, where $W$ collects
the sampled target matrices, and we keep all remaining parameters fixed.
For the diagnostic projector $P_\rho$, an ordinary gradient step in the
sampled fixed-basis factor coordinates induces $\Delta W=-\eta g_\rho$
in model weights.
For sufficiently small $\eta$,
\begin{equation}
\Delta\widetilde{\mathcal L}_r
=-\eta\langle g_r,g_\rho\rangle_F+O(\eta^2).
\label{eq:gradient_transfer}
\end{equation}
Positive alignment therefore identifies a shared local descent direction
within the sampled parameters. The pairwise mean alignments in
Figure~\ref{fig:lolift_mechanism}(b) are positive. At $\rho=0.6$,
cosine similarity is $0.59\pm0.02$ for $r=0.4$ and $0.56\pm0.11$
for $r=0.8$, with standard deviations across five disjoint batches.
This diagnostic tests local compatibility at initialization; the model
comparisons evaluate transfer after AdamW preparation and SVD compression
of the prepared model. Appendix~\ref{app:gradient_alignment} provides
the factor-coordinate derivation and probe details.
This fixed-projector diagnostic does not test whether the same local
benefit survives an AdamW update followed by SVD of the updated weights;
the following probe evaluates the combined effect of these two operations.

\paragraph{Realized update after recompression.}
We scale the first nonzero AdamW update from the default $\rho=0.6$ run
by $s$ and apply it to the original model, then recompute the deployment
SVD factors from the updated weights.
Let $\Delta J_r$ denote the change in held-out WT2 student-to-teacher KL
relative to compressing the original model at the same budget $r$.
KL decreases at all five tested budgets for
$s\in\{0.25,0.5,1\}$, with $\Delta J_r$ ranging from $-1.244$ to $-0.093$
nats/token at $s=1$ (Figure~\ref{fig:lolift_mechanism}(c)).
At $s=1$, updating only the target matrices yields nearly identical KL changes,
while the norm-matched random control stays near zero.
These measurements provide local evidence of cross-budget transfer
for the first nonzero update.
Appendix~\ref{app:gradient_alignment} reports the controls.

\subsection{Generality Across Compression Operators and Models}
\label{sec:transfer}
\label{sec:generality}

\begin{table}[!htbp]
\centering\footnotesize
\setlength{\tabcolsep}{1.5pt}
{\caption{Effect of LLift preparation across deployment compressors on LLaMA-7B. All methods use \texttt{mix75} calibration (75\% WT2 and 25\% Alpaca). All LLift rows use the same prepared checkpoint. Bold values mark the best result within each Original/LLift pair.}
\label{tab:compressor_transfer}}
\begin{tabular*}{\linewidth}{@{\extracolsep{\fill}}llrrrrrrrrrrrr@{}}
\toprule
 &  & \multicolumn{4}{c}{$r=0.8$} & \multicolumn{4}{c}{$r=0.6$} & \multicolumn{4}{c}{$r=0.4$} \\
\cmidrule(lr){3-6}\cmidrule(lr){7-10}\cmidrule(lr){11-14}
Compressor & Checkpoint & WT2 & PTB & C4 & Avg. & WT2 & PTB & C4 & Avg. & WT2 & PTB & C4 & Avg. \\
\midrule
SVD-LLM & Original & 8.03 & 14.03 & 13.89 & 0.44 & 14.27 & 38.34 & 35.03 & 0.40 & 63.25 & 300.26 & 218.52 & 0.31 \\
 & LLift & \textbf{7.49} & \textbf{11.72} & \textbf{11.15} & \textbf{0.50} & \textbf{8.40} & \textbf{15.96} & \textbf{14.77} & \textbf{0.46} & \textbf{13.48} & \textbf{53.47} & \textbf{32.24} & \textbf{0.38} \\
\midrule
ZS-SVD & Original & \textbf{6.86} & 11.42 & 10.46 & \textbf{0.51} & 12.41 & 34.31 & 28.04 & 0.42 & 56.71 & 291.02 & 184.77 & 0.33 \\
 & LLift & 7.84 & \textbf{11.39} & \textbf{10.11} & \textbf{0.51} & \textbf{9.86} & \textbf{19.22} & \textbf{15.38} & \textbf{0.45} & \textbf{16.88} & \textbf{88.31} & \textbf{35.33} & \textbf{0.38} \\
\midrule
LEMS & Original & \textbf{6.25} & \textbf{10.12} & \textbf{8.93} & \textbf{0.52} & 9.49 & 19.28 & 17.15 & 0.44 & 23.21 & 91.16 & 64.58 & 0.35 \\
 & LLift & 7.76 & 11.22 & 9.40 & \textbf{0.52} & \textbf{9.03} & \textbf{15.67} & \textbf{13.42} & \textbf{0.46} & \textbf{13.72} & \textbf{38.20} & \textbf{27.13} & \textbf{0.39} \\
\bottomrule
\end{tabular*}
\end{table}

\paragraph{Compression with different operators.}
Table~\ref{tab:compressor_transfer} tests whether LLift preparation
transfers across deployment operators. We compress the original
LLaMA-7B model and the same default LLift-prepared checkpoint with
SVD-LLM, one-shot ZS-SVD, and LEMS, using \texttt{mix75} calibration
and matched evaluation settings within each pair.
At $r=0.6/0.4$, preparation improves mean accuracy and lowers all
three perplexities for every compressor. With LEMS, mean accuracy
rises from $0.44/0.35$ to $0.46/0.39$, while C4 perplexity decreases
from $17.15/64.58$ to $13.42/27.13$.

At $r=0.8$, preparation leaves LEMS mean accuracy nearly unchanged
and increases all three perplexities. For ZS-SVD, WT2 perplexity
increases, whereas PTB and C4 perplexity decrease. The gains therefore
depend on both the compressor and deployment budget.
These comparisons show that one checkpoint prepared with
the SVD-LLM basis can also benefit a Fisher-based decomposition with
nonuniform rank allocation.

\begin{table*}[!htbp]
\centering

\begin{minipage}[t]{0.48\textwidth}
    \centering
    \caption{Evaluation across model families. 
    Seven-task average accuracy ($\uparrow$).
    }
    \label{tab:generality}

    \footnotesize
    \setlength{\tabcolsep}{2pt}
    \begin{tabular}{@{}llccc@{}}
    \toprule
Model & Method & $r{=}0.8$ & $r{=}0.6$ & $r{=}0.4$ \\
    \midrule
    Qwen3-8B
 & SVD-LLM & 0.49 & 0.35 & 0.31 \\
 & ZS-SVD & 0.55 & 0.41 & 0.31 \\
 & \textbf{LLift} & 0.55 & 0.47 & 0.34 \\
    \midrule
    Gemma-2-9B
 & SVD-LLM & 0.52 & 0.40 & 0.32 \\
 & ZS-SVD & 0.57 & 0.39 & 0.32 \\
 & \textbf{LLift} & 0.57 & 0.51 & 0.40 \\
    \bottomrule
    \end{tabular}
\end{minipage}
\hfill
\begin{minipage}[t]{0.48\textwidth}
    \centering
    \caption{Scaling to larger LLaMA models.
    Seven-task average accuracy ($\uparrow$).
    }
    \label{tab:scalability}

    \footnotesize
    \setlength{\tabcolsep}{2pt}
    \begin{tabular}{@{}llccc@{}}
    \toprule
Model & Method & $r{=}0.8$ & $r{=}0.6$ & $r{=}0.4$ \\
    \midrule
    LLaMA-13B
 & SVD-LLM & 0.50 & 0.42 & 0.32 \\
 & ZS-SVD & 0.54 & 0.39 & 0.34 \\
 & \textbf{LLift} & 0.54 & 0.49 & 0.41 \\
    \midrule
    LLaMA-30B
 & SVD-LLM & 0.55 & 0.45 & 0.33 \\
 & ZS-SVD & 0.59 & 0.52 & 0.38 \\
 & \textbf{LLift} & 0.57 & 0.53 & 0.45 \\
    \bottomrule
    \end{tabular}
\end{minipage}

\vspace{2pt}
\end{table*}

\paragraph{Model families and scales.}
Tables~\ref{tab:generality} and~\ref{tab:scalability} extend the evaluation to Qwen3-8B, Gemma-2-9B,
LLaMA-13B, and LLaMA-30B.
We prepare each model independently and use the same prepared model
for all three deployment budgets.
At $r=0.6/0.4$, LLift achieves higher mean accuracy than the
SVD-LLM and ZS-SVD references across all four models.
At $r=0.6$, accuracy on Qwen3-8B and Gemma-2-9B rises from
$0.35/0.40$ with SVD-LLM to $0.47/0.51$ with LLift.
At $r=0.4$, accuracy reaches $0.41$ on LLaMA-13B and
$0.45$ on LLaMA-30B, compared with $0.32$ and $0.33$
for SVD-LLM compression.
At $r=0.8$, LLift matches ZS-SVD's seven-task mean accuracy on
Qwen3-8B, Gemma-2-9B, and LLaMA-13B,
but trails it on LLaMA-30B ($0.57$ versus $0.59$).

\FloatBarrier
\section{Conclusion}

LLift prepares a reusable model for SVD compression across deployment budgets
by fine-tuning a low-rank view at a single preparation budget.
The frozen original model supervises the low-rank student.
We merge the learned updates into the original weights while preserving
components excluded from the student forward pass. We reuse SVDs of the
prepared model's transformed target matrices at different truncation ranks
to obtain deployment models without further fine-tuning.

On LLaMA-7B, a single prepared model raises seven-task mean accuracy
from the reported SVD-LLM scores of $0.44/0.37/0.31$ to $0.50/0.46/0.38$
at $r=0.8/0.6/0.4$. Controlled comparisons show that preserving the complementary 
components of the original weights is important for subsequent recompression, 
while the similar retained spectral energy before and after preparation indicates 
that spectral energy alone does not explain compression quality. 
The improvements further extend across model families, model sizes, 
and SVD-based compression operators. 
LLift thus shifts fine-tuning from individual compressed models to model
preparation, allowing the same prepared model to support multiple deployment
budgets without repeating compression-aware optimization.



\subsubsection*{Acknowledgments}
We are especially grateful to Ananda (Theertha) Suresh for guidance
throughout this project and for reviewing the manuscript.
We thank Amir Yazdanbakhsh for enabling this project and providing valuable advice.
We also thank Xin Lin and Henry Zheng for helpful discussions and suggestions.

\clearpage
\bibliography{iclr2027_conference}

\bibliographystyle{iclr2027_conference}

\clearpage
\appendix
\renewcommand{\topfraction}{0.95}
\renewcommand{\bottomfraction}{0.85}
\renewcommand{\textfraction}{0.05}
\renewcommand{\floatpagefraction}{0.8}
\setcounter{topnumber}{4}
\setcounter{totalnumber}{5}
\section{Additional Derivations}
\label{app:projection_derivation}

We prove the minimum-change completion and analyze reconstruction after
residual restoration. With fixed $Q$, we also show when training $B$
through $QB$ and training $W$ through $QQ^\top W$ produce identical
weight updates.

\subsection{Minimum-Change Completion}
\label{app:minimum_change}
We omit the layer index and let $Q_0^\top Q_0=I$,
$B_0=Q_0^\top W_0$, and $R_0=(I-Q_0Q_0^\top)W_0$.
Every feasible $W$ satisfying $Q_0^\top W=B^\star$ has the unique
decomposition $W=Q_0B^\star+R$ with $Q_0^\top R=0$.
The two terms of $W-W_0=Q_0(B^\star-B_0)+(R-R_0)$ are orthogonal in
the Frobenius inner product. Consequently,
\begin{equation}
\|W-W_0\|_F^2
=\|B^\star-B_0\|_F^2+\|R-R_0\|_F^2.
\label{eq:completion_pythagoras}
\end{equation}
The learned coordinates fix the first term, and $R=R_0$ uniquely minimizes
the second. This proves
Eq.~\ref{eq:minimum_change_completion} for each target matrix. The
minimization concerns target weights in Frobenius norm; we optimize the
remaining model parameters with the training objective in
Eq.~\ref{eq:LoLift_loss}.

\subsection{Residual Restoration and Reconstruction after Compression}
\label{app:residual_reconstruction}

\leavevmode{We compare compression with and without residual restoration using the
same deployment transform $T$ and rank $m$. Omitting the layer index, define
\[
H=Q_0B^\star T,\qquad
R=(W_0-Q_0B_0)T,\qquad
D=Q_0(B^\star-B_0)T.
\]
The teacher matrix is $A_0=W_0T=H+R-D$. Let $[X]_m$ denote a best
approximation to $X$ of rank at most $m$ in Frobenius norm. We compare
\[
Y_-=[H]_m,\qquad Y_+=[H+R]_m
\]
through their teacher reconstruction errors:
\[
\mathcal E(Y)=\|A_0-Y\|_F^2,\qquad
\Delta E_m=\mathcal E(Y_-)-\mathcal E(Y_+).
\]

Define
\[
\begin{aligned}
J_m&=\|H+R-Y_-\|_F^2-\|H+R-Y_+\|_F^2,\\
I_m&=2\langle D,Y_+-Y_-\rangle_F.
\end{aligned}
\]
Since $Y_+$ optimally approximates $H+R$ and $Y_-$ satisfies the same
rank constraint, $J_m\ge0$. Expanding
$\mathcal E(Y)=\|(H+R-Y)-D\|_F^2$ gives
\begin{equation}
\Delta E_m=J_m-I_m.
\label{eq:residual_gain_identity}
\end{equation}
Thus, restoration improves teacher reconstruction exactly when $J_m>I_m$.

Let $P$ be the orthogonal projector onto $\operatorname{col}(Q_0)$.
Since $PD=D$, Cauchy--Schwarz gives
\begin{equation}
\Delta E_m\ge
J_m-2\|D\|_F\|P(Y_+-Y_-)\|_F
=:M_m.
\label{eq:residual_sufficient_condition}
\end{equation}
Therefore, $M_m>0$ guarantees improvement at any deployment rank.
This bound does not require $Q_0^\top R_0=0$.

For deployment rank $m=k+t>k$, where $k$ is the preparation rank,
we have $Y_-=H$. The matrix $H+[R]_t$ has rank at most $m$, so
optimality of $Y_+$ implies
\[
\|F\|_F\le\|R-[R]_t\|_F,
\qquad F=H+R-Y_+.
\]
Because the two teacher reconstruction errors are $\|R-D\|_F^2$ and
$\|F-D\|_F^2$,
\[
\Delta E_m
=\|R\|_F^2-\|F\|_F^2
-2\langle R,D\rangle_F+2\langle F,D\rangle_F.
\]
Applying Cauchy--Schwarz and the SVD identity
$\|R\|_F^2-\|R-[R]_t\|_F^2=\|[R]_t\|_F^2$ yields
\begin{equation}
\Delta E_m\ge
\|[R]_t\|_F^2
-2\|D\|_F\|R-[R]_t\|_F
-2\langle R,D\rangle_F.
\label{eq:residual_extra_rank_bound}
\end{equation}
When $Q_0^\top R_0=0$, the final term vanishes. In this case,
\[
J_m\ge\|[R]_t\|_F^2,\qquad
\|P(Y_+-Y_-)\|_F=\|PF\|_F\le\|R-[R]_t\|_F,
\]
so Eq.~\ref{eq:residual_sufficient_condition} is at least as tight as
Eq.~\ref{eq:residual_extra_rank_bound}. The latter makes the contribution
of additional rank explicit through the residual spectrum. These bounds
characterize reconstruction of individual teacher weight matrices under
the fixed deployment transform.}

\subsection{Fixed-Basis Updates under SGD and AdamW}
For a fixed orthonormal basis $Q$ and projector $P=QQ^\top$, write
\begin{equation}
W=QB+R,\qquad B=Q^\top W,\qquad R=(I-P)W.
\label{eq:fixed_basis_decomposition}
\end{equation}
The full-matrix forward $PW$ equals the factorized forward $QB$.
For upstream gradient $G=\partial\mathcal L/\partial(PW)$, their
gradients satisfy $\nabla_W\mathcal L=Q Q^\top G$ and
$\nabla_B\mathcal L=Q^\top G$. Under plain SGD without momentum or weight
decay, an update to the full matrix therefore gives
\begin{equation}
W_{t+1}=W_t-\eta_t Q Q^\top G_t
       =Q\bigl(B_t-\eta_t Q^\top G_t\bigr)+R_0.
\label{eq:fixed_basis_sgd_equivalence}
\end{equation}
With matched initialization, minibatches, learning rates, and updates to
other parameters, the two implementations produce the same student
trajectory and the same prepared model with target weights $QB_*+R_0$.
For this fixed-basis SGD setting, factor coordinates therefore suffice
to represent the full-matrix training trajectory.

This equivalence does not generally extend to AdamW. Its elementwise second
moment statistics and adaptive scaling depend on whether the optimizer acts
in the coordinates of $W$ or $B$. Even when the gradient with respect to $W$ lies in the
column space of $Q$, the adaptively scaled direction $A_t$ need not do so.
For an AdamW update to $W$ with decay coefficient $\lambda$,
\begin{equation}
W_{t+1}=(1-\eta_t\lambda)W_t-\eta_t A_t,
\qquad
R_{t+1}=(1-\eta_t\lambda)R_t-\eta_t(I-P)A_t.
\label{eq:dense_adamw_residual}
\end{equation}
Thus full-matrix AdamW can update the complementary component, whereas
factorized training followed by residual restoration preserves $R_0$.
Refreshing the projector additionally changes the training subspace.
\section{Additional LLift Results and Evaluation Details}
\label{app:lolift_results}

We examine preparation choices, compression operators, and cross-budget
diagnostics. We report accuracy on a 0--1 scale and calculate differences
before rounding scores to two decimal places.

\subsection{Paired Diagnostic of Residual Restoration}
\label{app:residual_diagnostic}

\leavevmode{We use the LLaMA-7B checkpoint prepared at $\rho=0.6$
(Table~\ref{tab:controlled_sources}). We compress $Q_0B^\star$ and
$Q_0B^\star+R_0$ with identical adapted non-target parameters, frozen
\texttt{mix75} deployment transforms, and ranks at $r=0.8/0.6/0.4$.

For target matrix $\ell$ at budget $r$, we evaluate the quantities
$J_{\ell,r}$, $I_{\ell,r}$, and $M_{\ell,r}$ from
Appendix~\ref{app:residual_reconstruction} in FP64. We reconstruct
$R_0=W_0-Q_0B_0$ in FP64 and obtain $P$ by QR factorization. We separately
measure reconstruction gains using the exported compression factors:
\begin{equation}
\widehat{\Delta E}_{\ell,r}
=\|(W_{\ell,0}-\widehat W_{\ell,r}^{-})T_\ell\|_F^2
-\|(W_{\ell,0}-\widehat W_{\ell,r}^{+})T_\ell\|_F^2,
\label{eq:residual_measured_gain}
\end{equation}
where $\widehat W_{\ell,r}^{-}$ and $\widehat W_{\ell,r}^{+}$ denote
the compressed weights without and with restoration. We normalize each
gain by $N_\ell=\|W_{\ell,0}T_\ell\|_F^2$. The maximum normalized
discrepancy between the theoretical and exported reconstruction gains
is $1.14\times10^{-5}$.}

\begin{table}[!htbp]
\centering
\small
\setlength{\tabcolsep}{5pt}
\setlength{\belowcaptionskip}{5pt}
{\caption{Paired reconstruction diagnostic on LLaMA-7B. Positive
$M_{\ell,r}$ certifies improvement for the exact truncated matrices defined
in Appendix~\ref{app:residual_reconstruction}. The measured gain
$\widehat{\Delta E}_{\ell,r}$ uses exported compression factors.
Counts, means, and maxima are taken over 224 target matrices.}
\label{tab:residual_reconstruction}}
\begin{tabular*}{\linewidth}{@{\extracolsep{\fill}}crrrr@{}}
\toprule
$r$ & $M_{\ell,r}>0$ & $\widehat{\Delta E}_{\ell,r}>0$
& Mean $\widehat{\Delta E}_{\ell,r}/N_\ell$
& Max. $|I_{\ell,r}|/J_{\ell,r}$ \\
\midrule
0.8 & 224/224 & 224/224 & $3.882\times10^{-2}$ & $0.60\%$ \\
0.6 & 165/224 & 224/224 & $8.055\times10^{-3}$ & $1.93\%$ \\
0.4 & 179/224 & 224/224 & $5.948\times10^{-3}$ & $2.16\%$ \\
\bottomrule
\end{tabular*}
\end{table}

\leavevmode{The gain $J_{\ell,r}$ is strictly positive for every matrix at all three
budgets, and $|I_{\ell,r}|/J_{\ell,r}$ never exceeds $2.2\%$. Thus, the
interaction with the learned update does not offset the approximation
gain in this comparison. The measured teacher reconstruction error
decreases for all 224 matrices at each budget. The sufficient condition
certifies improvement for $224/165/179$ matrices at $r=0.8/0.6/0.4$,
while the remaining matrices also improve empirically. At $r=0.8$,
the bound expressed through the residual spectrum in
Eq.~\ref{eq:residual_extra_rank_bound} is positive for 216 matrices.

At $r=0.6/0.8$, the deployment rank is at least the preparation rank
for every matrix, so $[H]_m=H$ in exact arithmetic. Increasing the rank
beyond $k$ therefore retains no additional components of $H$, whereas
residual restoration makes additional components available to SVD.
Residual restoration also improves seven-task mean accuracy at all three
budgets (Table~\ref{tab:controlled_sources}).}

\subsection{Freezing the Remaining Parameters}
\label{app:dense_freeze}

To isolate the contribution of $\Phi$, we freeze the token embeddings,
output head, and all 65 RMSNorm modules in LLaMA-7B and train only the
factor coordinates $\mathcal B$. Both LLift variants use KL at
$\rho=0.6$ and restore the original residuals before compression,
with all other settings unchanged. Original receives no preparation.

\begin{table}[!htbp]
\centering
\small
\setlength{\tabcolsep}{8pt}
\setlength{\belowcaptionskip}{5pt}
\caption{Freezing non-target parameters $\Phi$ during LLaMA-7B preparation.
All checkpoints use SVD-LLM with the same fixed original-model
\texttt{mix75} activation transforms.
Avg. is seven-task mean plain accuracy.}
\label{tab:lolift_dense_freeze}
\begin{tabular}{@{}llrrr@{}}
\toprule
$r$ & Checkpoint & WT2 PPL $\downarrow$ & C4 PPL $\downarrow$ & Avg. $\uparrow$ \\
\midrule
0.8 & Original & 8.03 & 13.89 & 0.44 \\
 & LLift (frozen $\Phi$) & 8.40 & 11.58 & 0.50 \\
 & LLift (trainable $\Phi$) & 7.49 & 11.15 & 0.50 \\
\midrule
0.6 & Original & 14.27 & 35.03 & 0.40 \\
 & LLift (frozen $\Phi$) & 9.46 & 15.46 & 0.46 \\
 & LLift (trainable $\Phi$) & 8.40 & 14.77 & 0.46 \\
\midrule
0.4 & Original & 63.25 & 218.52 & 0.31 \\
 & LLift (frozen $\Phi$) & 15.12 & 33.65 & 0.38 \\
 & LLift (trainable $\Phi$) & 13.48 & 32.24 & 0.38 \\
\bottomrule
\end{tabular}
\end{table}

With $\Phi$ frozen, LLift reduces C4 perplexity relative to Original by
$16.6/55.9/84.6\%$ and raises mean accuracy from $0.44/0.40/0.31$
to $0.50/0.46/0.38$ at $r=0.8/0.6/0.4$
(Table~\ref{tab:lolift_dense_freeze}).
Compared with training $\Phi$, freezing it increases WT2 perplexity by
$12$--$13\%$ and C4 perplexity by approximately $4$--$5\%$ across
these budgets, while changing unrounded mean accuracy by at most $0.0022$.

\subsection{Objective and Anchor Ablations}
\label{app:objective_anchor}

\begin{figure}[!htbp]
    \centering
    \includegraphics[width=0.40\linewidth]{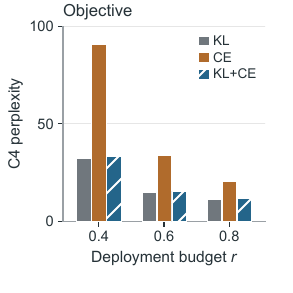}
    \caption{Training-objective ablation on LLaMA-7B. C4 perplexity for
    KL, CE, and equally weighted KL+CE at each deployment budget, with
    preparation anchor $\rho=0.6$.}
    \label{fig:lolift_objective}
\end{figure}

We vary the loss while keeping the remaining recipe fixed. KL$+$CE gives
each term weight one; pure CE uses no teacher.
Figure~\ref{fig:lolift_objective} shows that CE increases C4 perplexity
at all three budgets, from $32.2$ to $90.6$ at $r=0.4$. Its accuracy
slightly exceeds KL at $r=0.6$ but falls to $0.37$ at $r=0.4$, versus
$0.38$ for KL. KL$+$CE improves mean accuracy over KL by
$0.10$--$0.83$ percentage points while slightly increasing C4 perplexity
at every budget. We use KL to balance task accuracy and language-modeling
quality.

On LLaMA-7B, $\rho=0.6$ yields the lowest WT2 perplexity at all three
budgets among the tested anchors, while $\rho=0.4$ gives higher accuracy
at $r=0.6/0.4$ (Table~\ref{tab:lolift_anchor}).

\begin{table}[!htbp]
\centering
\small
\setlength{\tabcolsep}{3pt}
\setlength{\belowcaptionskip}{5pt}
{\caption{Preparation-anchor ablation on LLaMA-7B under SVD-LLM.
WT2 and C4 report perplexity ($\downarrow$); Avg. is seven-task mean
plain accuracy ($\uparrow$).}
\label{tab:lolift_anchor}}
\begin{tabular*}{\linewidth}{@{\extracolsep{\fill}}lrrrrrrrrr@{}}
\toprule
 & \multicolumn{3}{c}{$r=0.8$} & \multicolumn{3}{c}{$r=0.6$} & \multicolumn{3}{c}{$r=0.4$} \\
\cmidrule(lr){2-4}\cmidrule(lr){5-7}\cmidrule(lr){8-10}
Checkpoint & WT2 & C4 & Avg. & WT2 & C4 & Avg. & WT2 & C4 & Avg. \\
\midrule
Original & 8.03 & 13.89 & 0.44 & 14.27 & 35.03 & 0.40 & 63.25 & 218.52 & 0.31 \\
LLift ($\rho=0.4$) & 9.72 & 12.78 & 0.50 & 10.89 & 17.08 & 0.47 & 14.07 & 30.78 & 0.39 \\
LLift ($\rho=0.6$) & 7.49 & 11.15 & 0.50 & 8.40 & 14.77 & 0.46 & 13.48 & 32.24 & 0.38 \\
LLift ($\rho=0.8$) & 7.70 & 11.22 & 0.50 & 9.53 & 16.91 & 0.45 & 20.20 & 52.31 & 0.35 \\
\bottomrule
\end{tabular*}
\end{table}

\paragraph{Anchor selection beyond LLaMA-7B.}
\label{app:qwen3_anchor}
\leavevmode{We evaluate Qwen3-8B at $\rho\in\{0.4,0.6,0.8\}$ using seed 42,
with the checkpoint from the main experiments used for $\rho=0.6$.
We compress each checkpoint
at $r\in\{0.8,0.6,0.4\}$ with SVD-LLM and fixed original-model
\texttt{mix75} activation transforms (Table~\ref{tab:qwen3_anchor}).}

\begin{table}[!htbp]
\centering
\small
\setlength{\tabcolsep}{3pt}
\setlength{\belowcaptionskip}{5pt}
{\caption{Preparation-anchor ablation on Qwen3-8B under SVD-LLM.
WT2 and C4 report perplexity ($\downarrow$); Avg. is seven-task mean
plain accuracy ($\uparrow$).}
\label{tab:qwen3_anchor}}
\begin{tabular*}{\linewidth}{@{\extracolsep{\fill}}lrrrrrrrrr@{}}
\toprule
 & \multicolumn{3}{c}{$r=0.8$} & \multicolumn{3}{c}{$r=0.6$} & \multicolumn{3}{c}{$r=0.4$} \\
\cmidrule(lr){2-4}\cmidrule(lr){5-7}\cmidrule(lr){8-10}
Checkpoint & WT2 & C4 & Avg. & WT2 & C4 & Avg. & WT2 & C4 & Avg. \\
\midrule
Original & 14.21 & 28.97 & 0.52 & 31.79 & 89.05 & 0.38 & 239.87 & 896.79 & 0.31 \\
LLift ($\rho=0.4$) & 17.95 & 37.03 & 0.54 & 21.09 & 54.99 & 0.47 & 32.40 & 120.76 & 0.37 \\
LLift ($\rho=0.6$) & 13.95 & 29.55 & 0.55 & 17.85 & 46.78 & 0.47 & 44.60 & 173.05 & 0.34 \\
LLift ($\rho=0.8$) & 12.66 & 26.23 & 0.54 & 18.76 & 51.12 & 0.43 & 93.18 & 434.36 & 0.32 \\
\bottomrule
\end{tabular*}
\end{table}

\leavevmode{The default $\rho=0.6$ improves mean accuracy over direct compression
by $2.41/9.28/3.39$ percentage points at $r=0.8/0.6/0.4$.
Among the tested anchors, $\rho=0.4$ gives the highest accuracy and
lowest WT2/C4 perplexity at $r=0.4$, whereas $\rho=0.8$ gives the
lowest perplexity at $r=0.8$. The preferred anchor therefore depends
on the deployment budget and metric.

Across LLaMA-7B and Qwen3-8B, the tested anchors $\rho=0.4$ and $0.6$
generally provide a favorable balance of task accuracy and perplexity
over the evaluated deployment budgets. These results support using
$0.4$--$0.6$ as a practical anchor range for cross-budget preparation.}

\subsection{Preparation Bases and Compression Operators}
\label{app:cross_operator_details}

Table~\ref{tab:lolift_cross_tasks_zs} groups the preparation-basis
comparison by deployment operator. We prepare each LLift model once
at $\rho=0.6$ using either an SVD-LLM or a ZS-SVD basis, then compress
it with each operator at three budgets. Both deployment protocols use
\texttt{mix75} calibration.
Each deployment block evaluates the two preparation bases under the same
compression operator and evaluation settings.

\begin{table}[!htbp]
\footnotesize
\setlength{\tabcolsep}{2pt}
\caption{LLaMA-7B results across preparation bases and compression operators under \texttt{mix75} calibration. Parentheses identify the preparation basis; both LLift variants use KL at $\rho=0.6$. WT2 and C4 report perplexity; task columns and Avg. report zero-shot accuracy. Each panel reports one deployment operator.}
\label{tab:lolift_cross_tasks_zs}
\begin{center}
\begin{tabular*}{\linewidth}{@{\extracolsep{\fill}}llrrrrrrrrrr@{}}
\toprule
Preparation & $r$ & WT2 & C4 & OBQA & ARC-e & ARC-c & WinoG & HellaS & PIQA & MathQA & Avg. \\
\midrule
\multicolumn{12}{l}{SVD-LLM deployment} \\
\midrule
Original & 0.8 & 8.03 & 13.89 & 0.23 & 0.62 & 0.31 & 0.63 & 0.39 & 0.69 & 0.23 & 0.44 \\
 & 0.6 & 14.27 & 35.03 & 0.24 & 0.53 & 0.26 & 0.57 & 0.34 & 0.64 & 0.22 & 0.40 \\
 & 0.4 & 63.25 & 218.52 & 0.15 & 0.28 & 0.21 & 0.49 & 0.27 & 0.55 & 0.22 & 0.31 \\
\midrule
LLift (SVD-LLM) & 0.8 & 7.49 & 11.15 & 0.30 & 0.67 & 0.37 & 0.66 & 0.52 & 0.74 & 0.25 & 0.50 \\
 & 0.6 & 8.40 & 14.77 & 0.25 & 0.63 & 0.31 & 0.62 & 0.45 & 0.71 & 0.25 & 0.46 \\
 & 0.4 & 13.48 & 32.24 & 0.17 & 0.48 & 0.24 & 0.57 & 0.34 & 0.62 & 0.22 & 0.38 \\
\midrule
LLift (ZS-SVD) & 0.8 & 7.74 & 11.61 & 0.27 & 0.67 & 0.34 & 0.66 & 0.51 & 0.74 & 0.26 & 0.50 \\
 & 0.6 & 9.41 & 17.05 & 0.24 & 0.63 & 0.29 & 0.63 & 0.44 & 0.70 & 0.24 & 0.45 \\
 & 0.4 & 17.96 & 52.73 & 0.17 & 0.42 & 0.21 & 0.55 & 0.32 & 0.60 & 0.22 & 0.36 \\
\midrule
\multicolumn{12}{l}{ZS-SVD deployment} \\
\midrule
Original & 0.8 & 6.86 & 10.46 & 0.30 & 0.71 & 0.37 & 0.69 & 0.50 & 0.75 & 0.25 & 0.51 \\
 & 0.6 & 12.41 & 28.04 & 0.23 & 0.59 & 0.29 & 0.62 & 0.37 & 0.67 & 0.22 & 0.42 \\
 & 0.4 & 56.71 & 184.77 & 0.15 & 0.34 & 0.21 & 0.53 & 0.28 & 0.56 & 0.21 & 0.33 \\
\midrule
LLift (SVD-LLM) & 0.8 & 7.84 & 10.11 & 0.30 & 0.70 & 0.39 & 0.66 & 0.54 & 0.76 & 0.25 & 0.51 \\
 & 0.6 & 9.86 & 15.38 & 0.23 & 0.62 & 0.30 & 0.61 & 0.44 & 0.69 & 0.23 & 0.45 \\
 & 0.4 & 16.88 & 35.33 & 0.17 & 0.46 & 0.25 & 0.58 & 0.34 & 0.61 & 0.23 & 0.38 \\
\midrule
LLift (ZS-SVD) & 0.8 & 7.27 & 10.02 & 0.29 & 0.71 & 0.37 & 0.66 & 0.54 & 0.76 & 0.26 & 0.51 \\
 & 0.6 & 8.58 & 14.34 & 0.25 & 0.65 & 0.31 & 0.62 & 0.45 & 0.70 & 0.23 & 0.46 \\
 & 0.4 & 16.38 & 42.18 & 0.16 & 0.43 & 0.21 & 0.55 & 0.33 & 0.60 & 0.22 & 0.36 \\
\bottomrule
\end{tabular*}
\end{center}
\end{table}

Relative to the default SVD-LLM basis, the ZS-SVD basis changes accuracy
by $-0.11/+1.29/-1.75$ percentage points under ZS-SVD deployment at
$r=0.8/0.6/0.4$, and by $-0.60/-0.82/-2.32$ points under SVD-LLM
deployment. Matching the preparation and deployment operators therefore
does not uniformly improve transfer.

\leavevmode{In Table~\ref{tab:compressor_transfer}, SVD-LLM uses fixed original-model
activation transforms. For LEMS, we compute KFAC-SVD statistics and
select ranks for each checkpoint using the same \texttt{mix75} windows.
We compare the original and prepared checkpoints within each compressor
under matched evaluation settings.}

\paragraph{Deployment recalibration.}
\label{app:deployment_recalibration}
We re-estimate activation statistics from the LLift-prepared model,
apply SVD-LLM to its weights using the resulting transforms, and truncate
the shared decompositions at $r\in\{0.4,0.5,0.6,0.7,0.8\}$.
Table~\ref{tab:lolift_recalibration} reports changes relative to
compressing the same weights with original-model statistics.

\begin{table}[!htbp]
\small
\setlength{\tabcolsep}{8pt}
\caption{Deployment recalibration of LLaMA-7B LLift under SVD-LLM. Entries are paired differences (prepared-model statistics minus original-model statistics) for the same weights. Accuracy differences use the 0--1 scale; PPL differences are absolute.}
\label{tab:lolift_recalibration}
\begin{center}
\begin{tabular}{@{}lrrrrr@{}}
\toprule
Metric & $r=0.4$ & $r=0.5$ & $r=0.6$ & $r=0.7$ & $r=0.8$ \\
\midrule
$\Delta$ WT2 PPL & $+1.17$ & $+0.83$ & $+0.61$ & $+0.44$ & $+0.36$ \\
$\Delta$ C4 PPL & $-0.06$ & $+0.31$ & $+0.30$ & $+0.26$ & $+0.19$ \\
$\Delta$ Avg. accuracy & $-0.0021$ & $+0.0073$ & $+0.0035$ & $+0.0001$ & $+0.0008$ \\
\bottomrule
\end{tabular}
\end{center}
\end{table}

Recalibration changes mean accuracy by $-0.21$ to $+0.73$ percentage
points. WT2 perplexity increases by $0.36$--$1.17$ at every
budget; C4 perplexity changes by $-0.06$ to $+0.31$ and increases at
four of the five budgets. We retain original-model statistics in the
main comparisons because recalibration provides no consistent
perplexity benefit.

\subsection{Retained Spectral Energy}
\label{app:lolift_mechanism}

We apply SVD to each $W_\ell T_\ell$ using the same frozen
original-model \texttt{mix75} transforms and select prefixes according
to Eq.~\ref{eq:rank_budget}. We measure retained whitened spectral
energy as the fraction of squared singular values that remain after
truncation. Table~\ref{tab:lolift_energy} shows that the reported retained
fractions differ by at most $2\times10^{-4}$ despite the improvements
in compressed-model quality.

\begin{table}[!htbp]
\small
\setlength{\tabcolsep}{8pt}
\caption{Retained whitened spectral energy after truncation at each deployment budget. Both models use the same frozen \texttt{mix75} activation transforms.}
\label{tab:lolift_energy}
\begin{center}
\begin{tabular}{@{}lrrr@{}}
\toprule
Model & $r=0.4$ & $r=0.6$ & $r=0.8$ \\
\midrule
Original & 0.8818 & 0.9325 & 0.9633 \\
LLift & 0.8820 & 0.9327 & 0.9634 \\
\bottomrule
\end{tabular}
\end{center}
\end{table}

\subsection{Gradient Alignment and Recompression Probes}
\label{app:gradient_alignment}

We examine cross-budget transfer through gradient alignment at
initialization and through the measured effect of an AdamW update
followed by SVD recompression.

\begin{figure}[!htbp]
    \centering
    \includegraphics[width=.72\linewidth]{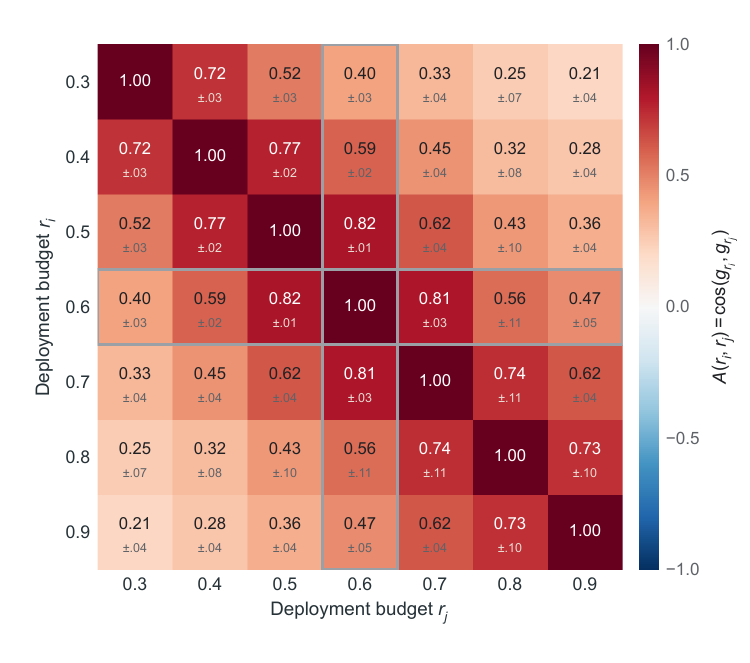}
    \caption{Cross-budget gradient alignment at initialization.
    Cells report mean cosine similarity and standard deviation across five
    disjoint WT2 batches. Figure~\ref{fig:lolift_mechanism}(b)
    shows the $0.4$--$0.8$ submatrix of means.}
    \label{fig:gradient_alignment_full}
\end{figure}

We use SVD-LLM projections and student-to-teacher KL from
Eq.~\ref{eq:LoLift_loss}, with temperature one.
We sample 56 of 224 target matrices across eight depths and use five
disjoint WT2 batches of $32\times1024$ input positions.
We evaluate all budgets at the original model and hold activation
statistics, each budget's projector, and the remaining parameters fixed
during differentiation. Let $W$ collect the sampled target matrices and define
\begin{equation}
g_r=\left.\nabla_W\widetilde{\mathcal L}_r(W;P_r)\right|_{W=W_0},
\qquad A(\rho,r)=\frac{\langle g_\rho,g_r\rangle_F}
{\|g_\rho\|_F\|g_r\|_F}.
\label{eq:gradient_alignment}
\end{equation}
Here $\widetilde{\mathcal L}_r$ is the fixed-projector KL surrogate, and
inner products sum over the sampled matrices.

For an orthonormal basis $Q_\rho$ of the diagnostic projector
$P_\rho=Q_\rho Q_\rho^\top$ and upstream gradient $G_\rho$, the
corresponding factor coordinates satisfy
$\nabla_B\mathcal L_\rho=Q_\rho^\top G_\rho$.
An ordinary gradient step confined to the sampled factor coordinates
therefore induces the model-weight update
$\Delta W=Q_\rho\Delta B=-\eta Q_\rho Q_\rho^\top G_\rho=-\eta g_\rho$.
A first-order expansion of $\widetilde{\mathcal L}_r$ along this update
gives Eq.~\ref{eq:gradient_transfer} for sufficiently small $\eta$.
All off-diagonal means in Figure~\ref{fig:gradient_alignment_full} are
positive, identifying a shared local descent component at initialization.
Standard deviations measure variation across probe batches at the same
model checkpoint.

\paragraph{Recompression probe and controls.}
For the update probe in Figure~\ref{fig:lolift_mechanism}(c), let
$\Delta\Theta$ denote the first nonzero AdamW displacement from the
default $\rho=0.6$ run. We measure
$J_r(\Theta)=\mathrm{KL}(p_{\mathrm S,r}(\Theta)\|p_{\mathrm T})$
on 64 held-out WT2 validation sequences of length 1024, with
no overlap with training batches, at temperature one. Here
$p_{\mathrm S,r}(\Theta)$ comes from compressing $\Theta$ at budget
$r$ and $p_{\mathrm T}$ comes from the frozen original teacher.
For each scale $s$, we recompute the weighted SVDs of the updated target
matrices using the frozen original-model \texttt{mix75} deployment
transforms and report
\[
\Delta J_r(s)=J_r(\Theta_0+s\Delta\Theta)-J_r(\Theta_0).
\]
Table~\ref{tab:adamw_recompression} reports the full update at
$s\in\{0.25,0.5,1\}$ and the target-only and norm-matched random
controls at $s=1$; negative values indicate improvement.
To isolate the target-matrix contribution, we write
$\Delta\Theta=(\Delta W,\Delta\Phi)$ and repeat the probe with
$(\Delta W,0)$, keeping the remaining parameters at $\Phi_0$.

\begin{table}[!htbp]
\small
\setlength{\tabcolsep}{5pt}
\caption{Change in held-out student-to-teacher KL after scaling the first nonzero AdamW update and recomputing the deployment SVD factors. Entries report $\Delta J_r(s)$ in $10^{-3}$ nats/token; negative values indicate improvement over compressing the original model. Signed zeros reflect rounding.}
\label{tab:adamw_recompression}
\begin{center}
\begin{tabular}{@{}lrrrrrr@{}}
\toprule
Update & $s$ & $r=0.4$ & $r=0.5$ & $r=0.6$ & $r=0.7$ & $r=0.8$ \\
\midrule
Full $(\Delta W,\Delta\Phi)$ & 0.25 & $-570$ & $-508$ & $-283$ & $-120$ & $-53$ \\
 & 0.5 & $-932$ & $-793$ & $-412$ & $-177$ & $-79$ \\
 & 1 & $-1244$ & $-1007$ & $-504$ & $-218$ & $-93$ \\
\midrule
Target only $(\Delta W,0)$ & 1 & $-1223$ & $-994$ & $-497$ & $-213$ & $-91$ \\
Norm-matched random & 1 & $+0.8$ & $+0.4$ & $+0.2$ & $-0.0$ & $+0.0$ \\
\bottomrule
\end{tabular}
\end{center}
\end{table}

At $s=1$, the target-only changes match the full-update changes within
$2.3\%$, while the random control stays near zero.
For this first nonzero update at $s\in\{0.25,0.5,1\}$, we also compare
KL changes under fixed projectors with those after recomputing the
deployment SVD factors. The changes have the same sign; the largest
reported relative difference is about $37\%$, at $s=0.25$ and $r=0.4$.

\subsection{Budget-Specific Fine-Tuning}
\label{app:budget_finetuning}

We compress the original LLaMA-7B model with SVD-LLM using
WT2 calibration, then fine-tune each compressed model independently
on Alpaca. We first train LoRA adapters on the \texttt{\detokenize{*_u_proj}}
factors and merge them, then train and merge adapters on the
\texttt{\detokenize{*_v_proj}} factors. We use a fixed configuration with
seed 42, three epochs per stage, and a learning rate of $10^{-4}$.
We retain the weights from the final epoch of each stage, without model
selection.
We evaluate mean plain accuracy over the same seven tasks.

\begin{table}[!htbp]
\small
\setlength{\tabcolsep}{5pt}
\caption{Budget-specific fine-tuning and LLift on LLaMA-7B. SVD-LLM + FT uses WT2 calibration followed by two-stage LoRA fine-tuning on Alpaca at each budget; Before FT reports the accuracy of its compressed initialization. LLift uses one model prepared at $\rho=0.6$ and frozen \texttt{mix75} deployment transforms for all budgets. WT2 and C4 report perplexity; Avg. is seven-task mean plain accuracy.}
\label{tab:budget_finetuning}
\begin{center}
\begin{tabular*}{\linewidth}{@{\extracolsep{\fill}}rrrrrrrr@{}}
\toprule
$r$ & \shortstack{Before FT\\Avg.} & \multicolumn{3}{c}{SVD-LLM + FT} & \multicolumn{3}{c}{LLift} \\
\cmidrule(lr){3-5}\cmidrule(lr){6-8}
 &  & WT2 $\downarrow$ & C4 $\downarrow$ & Avg. $\uparrow$ & WT2 $\downarrow$ & C4 $\downarrow$ & Avg. $\uparrow$ \\
\midrule
0.8 & 0.38 & 7.62 & 12.01 & 0.50 & 7.49 & 11.15 & 0.50 \\
0.6 & 0.35 & 9.54 & 15.78 & 0.46 & 8.40 & 14.77 & 0.46 \\
0.4 & 0.31 & 15.19 & 26.79 & 0.39 & 13.48 & 32.24 & 0.38 \\
\bottomrule
\end{tabular*}
\end{center}
\end{table}

\begin{samepage}
Table~\ref{tab:budget_finetuning} compares the resulting models with the
single LLift model prepared at $\rho=0.6$. At $r=0.8/0.6/0.4$, LLift's
accuracy is $0.15/0.22/1.13$ percentage points lower, while its WT2
perplexity is lower at all three budgets. LLift also has lower C4 perplexity
at $r=0.8/0.6$, while budget-specific fine-tuning has lower C4 perplexity
at $r=0.4$.
\par
\end{samepage}

\paragraph{{Matched recovery.}}
\label{app:matched_recovery}
\leavevmode{We apply SVD-LLM to the original checkpoint at each budget using
the same frozen WT2/Alpaca 75:25 calibration transforms and rank rule as
LLift. We then recover each compressed model with LLift's WT2 KL training
recipe, training the right factors and non-target parameters while
freezing the left factors. Each recovery run receives the same training
exposure as LLift: approximately $11.5$M token positions for LLaMA-7B
and $10.2$M for Qwen3-8B. All runs use seed 42.}

\begin{table}[!htbp]
\centering\footnotesize{}
\setlength{\tabcolsep}{3pt}
\setlength{\belowcaptionskip}{5pt}
{\caption{Budget-specific SVD-LLM recovery and LLift.
WT2 and C4 report perplexity ($\downarrow$).
Avg. is seven-task mean plain accuracy ($\uparrow$).}
\label{tab:matched_recovery}}
\begin{tabular*}{\linewidth}{@{\extracolsep{\fill}}l*{9}{r}@{}}
\toprule
 & \multicolumn{3}{c}{$r=0.8$} & \multicolumn{3}{c}{$r=0.6$} & \multicolumn{3}{c}{$r=0.4$} \\
\cmidrule(lr){2-4}\cmidrule(lr){5-7}\cmidrule(lr){8-10}
Method & WT2 & C4 & Avg. & WT2 & C4 & Avg. & WT2 & C4 & Avg. \\
\midrule
\multicolumn{10}{l}{LLaMA-7B} \\
LLift & 7.49 & 11.15 & 0.50 & 8.40 & 14.77 & 0.46 & 13.48 & 32.24 & 0.38 \\
Recovery & 6.75 & 10.52 & 0.50 & 7.96 & 14.05 & 0.46 & 10.97 & 23.40 & 0.40 \\
\midrule
\multicolumn{10}{l}{Qwen3-8B} \\
LLift & 13.95 & 29.55 & 0.55 & 17.85 & 46.78 & 0.47 & 44.60 & 173.05 & 0.34 \\
Recovery & 12.08 & 28.45 & 0.55 & 15.78 & 38.82 & 0.48 & 26.26 & 79.88 & 0.39 \\
\bottomrule
\end{tabular*}
\end{table}

\leavevmode{Table~\ref{tab:matched_recovery} compares three independently
recovered models with one LLift-prepared checkpoint. Mean accuracy is
broadly comparable at $r=0.8/0.6$, while recovery achieves higher accuracy
at $r=0.4$ and lower perplexity across all three budgets.}

\leavevmode{Table~\ref{tab:matched_recovery_gsm8k} reports Qwen3-8B GSM8K
results under the evaluation protocol in Appendix~\ref{app:gsm8k}.
Per-budget exposure assigns one LLift run's training exposure to each
recovery model. Total exposure divides that exposure across $N=3$ budgets
($r=0.8/0.6/0.4$) or $N=5$ budgets ($r=0.8/0.7/0.6/0.5/0.4$).
LLift uses the same prepared checkpoint throughout.}

\begin{table}[!htbp]
\centering\small{}
\setlength{\tabcolsep}{4pt}
\setlength{\belowcaptionskip}{5pt}
{\caption{Qwen3-8B GSM8K strict exact-match accuracy
(\%, $\uparrow$).}
\label{tab:matched_recovery_gsm8k}}
\begin{tabular*}{\linewidth}{@{\extracolsep{\fill}}lrrrrr@{}}
\toprule
Method & $r=0.8$ & $r=0.7$ & $r=0.6$ & $r=0.5$ & $r=0.4$ \\
\midrule
LLift & 64.67 & 46.32 & 17.13 & 2.05 & 0.38 \\
Recovery, per-budget ($N=3$) & 61.41 & --- & 17.97 & --- & 1.21 \\
Recovery, total ($N=3$) & 60.65 & --- & 13.65 & --- & 1.44 \\
Recovery, total ($N=5$) & 58.15 & 22.82 & 6.97 & 2.35 & 0.76 \\
\bottomrule
\end{tabular*}
\end{table}

\subsection{Preparation and Baseline Protocols}
\label{app:baseline_protocols}

\paragraph{Target matrices and preparation.}
We factorize the attention projections (\texttt{q\_proj},
\texttt{k\_proj}, \texttt{v\_proj}, \texttt{o\_proj}) and MLP
projections (\texttt{gate\_proj}, \texttt{up\_proj},
\texttt{down\_proj}) in every Transformer block. LLaMA-7B has 224 such
matrices across 32 blocks. We group the token embeddings, output head,
and normalization parameters into $\Phi$ and train them without
factorization. For the LLaMA-7B reference, we use seed 42, teacher KL
with temperature one, and the WT2 training split. We run AdamW
for four epochs (176 optimizer steps) with learning rate $10^{-4}$,
warmup fraction $0.05$, weight decay $0.01$, gradient clipping at norm
1, sequence length 1024, and global batch size 64. We construct the
preparation bases from original-model activations on WT2 and
freeze both the teacher and these bases.

\paragraph{Deployment calibration and scoring.}
We estimate activation transforms from the original model on 256
sequences of length 2048. This calibration set contains
75\% WT2 and 25\% Alpaca (\texttt{mix75}); we construct it
separately from preparation calibration. We freeze these transforms
across models and budgets, decompose each $W_\ell T_\ell$ once, and
select prefixes for each deployment budget. We evaluate WT2
perplexity on its held-out test split and C4 perplexity separately.
We average unrounded plain accuracies equally over the seven tasks in
Sec.~\ref{sec:experiments}. The paired residual-restoration control keeps
the adapted non-target parameters and deployment transforms identical.

\paragraph{Compression and elastic baselines.}
We use the SVD baselines documented by
\citet{abbasi2026zerosumsvdbalancing,hu2026saessvdselfadaptivesuppressionaccumulated}.
For cross-model experiments, SVD-LLM uses 256
WT2 sequences of length 2048, and ZS-SVD uses 256 WT2
samples.

\leavevmode{We run Swift-SVD$^*$ with the released dynamic-rank configuration
($\alpha=0.6$) and C4 calibration, and LEMS with WT2 calibration.
Both use 256 sequences of length 2048. LEMS uses 32 rank-search
samples and the HiGHS solver in place of Gurobi.
Our LLaMA-7B Swift-SVD reproduction falls below the paper's reported accuracies.
Both methods select ranks for each target budget without
budget-specific model-weight training.}

For the elastic comparison, ACIP uses C4 and FlexRank uses FineWeb-Edu
for preparation, while LLift uses WT2. ACIP omits the rank-32
correction $\Delta$; ACIP ($+\Delta$) retains it. The reported budget
for ACIP ($+\Delta$) excludes this correction's $79.95$M parameters.

\paragraph{Spectral-regularization control.}
\label{app:spectral_control}
For the spectral control in Table~\ref{tab:controlled_sources}, we train
the dense LLaMA-7B student with student-to-teacher KL at temperature one,
without CE or a low-rank forward pass. We match LLift's AdamW recipe
and 11.5M-token training budget described above.
For each of the 224 attention and MLP target matrices
$W_\ell\in\mathbb R^{m_\ell\times n_\ell}$, we penalize the normalized
truncated Schatten-1 tail
\begin{equation}
\mathcal R_\ell(W_\ell)
=\frac{\sum_{i>k_\ell}\sigma_i(W_\ell T_\ell)}
{\|W_\ell T_\ell\|_F},
\qquad
k_\ell=\left\lfloor\frac{0.6\,m_\ell n_\ell}{m_\ell+n_\ell}\right\rfloor.
\label{eq:spectral_control}
\end{equation}
We compute and freeze the whitening factors $T_\ell$ from the original
model on 256 WT2 sequences of length 1024. We compute spectra
from FP32 Gram eigenvalues floored at $10^{-6}$.
At each optimizer step, we apply a separate Muon update for
the spectral term with coefficient $10^{-3}$, learning rate $10^{-5}$,
momentum $0.95$, five Newton--Schulz iterations, and RMS scale $0.2$.
We compress the trained weights with SVD-LLM using frozen original-model
\texttt{mix75} transforms and the shared evaluation protocol described above.

\subsection{Additional Comparisons on Qwen3-8B}
\label{app:qwen3_additional_baselines}

\leavevmode{Table~\ref{tab:qwen3_additional_baselines} compares SVD-LLM, one-shot
ZS-SVD, the released Swift-SVD$^*$ configuration~\citep{qi2026swiftsvd},
and KFAC-SVD with LEMS rank allocation~\citep{thoma2026lems}, denoted LEMS,
on Qwen3-8B. We use the shared evaluation protocol to evaluate
SVD-LLM perplexity and ZS-SVD.
\mbox{LLift + SVD-LLM} and \mbox{LLift + LEMS} apply SVD-LLM and KFAC-SVD with
LEMS rank allocation, respectively, to the same checkpoint prepared at $\rho=0.6$.}

\begin{table}[!htbp]
\centering
\small
\setlength{\tabcolsep}{3pt}
\setlength{\belowcaptionskip}{5pt}
{\caption{Additional SVD compression results on Qwen3-8B.
WT2 and C4 report perplexity ($\downarrow$); Avg. is seven-task mean
plain accuracy ($\uparrow$).}
\label{tab:qwen3_additional_baselines}}
\begin{tabular*}{\linewidth}{@{\extracolsep{\fill}}lrrrrrrrrr@{}}
\toprule
 & \multicolumn{3}{c}{$r=0.8$} & \multicolumn{3}{c}{$r=0.6$} & \multicolumn{3}{c}{$r=0.4$} \\
\cmidrule(lr){2-4}\cmidrule(lr){5-7}\cmidrule(lr){8-10}
Method & WT2 & C4 & Avg. & WT2 & C4 & Avg. & WT2 & C4 & Avg. \\
\midrule
SVD-LLM & 14.06 & 37.63 & 0.49 & 28.96 & 136.92 & 0.35 & 186.53 & 1344.52 & 0.31 \\
ZS-SVD & 12.53 & 27.63 & 0.55 & 61.29 & 213.74 & 0.40 & 165.57 & 720.59 & 0.31 \\
Swift-SVD$^*$ & 23.57 & 21.97 & 0.53 & 66.82 & 41.62 & 0.43 & 301.21 & 130.99 & 0.33 \\
LEMS & 10.26 & 18.01 & 0.56 & 15.04 & 54.35 & 0.43 & 40.48 & 222.46 & 0.33 \\
LLift + SVD-LLM & 13.95 & 29.55 & 0.55 & 17.85 & 46.78 & 0.47 & 44.60 & 173.05 & 0.34 \\
LLift + LEMS & 12.71 & 21.44 & 0.59 & 15.24 & 29.75 & 0.51 & 26.68 & 71.90 & 0.38 \\
\bottomrule
\end{tabular*}
\end{table}

\leavevmode{\mbox{LLift + SVD-LLM} achieves higher mean accuracy than SVD-LLM and
Swift-SVD$^*$ at all three budgets and than ZS-SVD and LEMS at
$r=0.6/0.4$, while LEMS performs better
at $r=0.8$. Deploying the same prepared checkpoint with LEMS yields
accuracies of $0.59/0.51/0.38$ at $r=0.8/0.6/0.4$ and lower
WT2/C4 perplexity than \mbox{LLift + SVD-LLM} at every
budget. This composition requires no further model training.}

\subsection{Deployment Calibration Sensitivity}
\label{app:calibration_sensitivity}

\leavevmode{We keep the LLaMA-7B LLift checkpoint fixed and compare deployment
calibration on WT2, C4, and the default WT2/Alpaca 75:25 mixture
(\texttt{mix75}), using 256 sequences of length 2048 per corpus.
For each corpus, we apply SVD-LLM to both checkpoints with shared
original-model activation transforms.
Table~\ref{tab:calibration_sensitivity} shows seven-task mean accuracy
ranges of $0.49$--$0.50$, $0.45$--$0.46$, and $0.35$--$0.38$ for
LLift at $r=0.8/0.6/0.4$, respectively.
At $r=0.4$, LLift achieves $0.35/0.36/0.38$ with WT2/C4/\texttt{mix75},
compared with $0.31/0.30/0.31$ for the original checkpoint under
matched calibration. Preparation gains therefore persist with WT2
and C4, while \texttt{mix75} further improves LLift's accuracy.}

\begin{table}[!htbp]
\centering
\small
\setlength{\tabcolsep}{5pt}
\setlength{\belowcaptionskip}{5pt}
{\caption{Deployment calibration sensitivity on LLaMA-7B under SVD-LLM.
Original and LLift share original-model activation transforms within each
calibration setting. WT2 and C4 report perplexity; Avg. is seven-task
mean plain accuracy.}
\label{tab:calibration_sensitivity}}
\begin{tabular*}{\linewidth}{@{\extracolsep{\fill}}llrrrrrr@{}}
\toprule
 & & \multicolumn{2}{c}{WT2 PPL $\downarrow$} & \multicolumn{2}{c}{C4 PPL $\downarrow$} & \multicolumn{2}{c}{Avg. $\uparrow$} \\
\cmidrule(lr){3-4}\cmidrule(lr){5-6}\cmidrule(lr){7-8}
$r$ & Calibration & Original & LLift & Original & LLift & Original & LLift \\
\midrule
0.8 & WT2 & 7.90 & 7.14 & 15.85 & 11.46 & 0.38 & 0.50 \\
 & C4 & 12.01 & 12.02 & 12.01 & 10.66 & 0.46 & 0.49 \\
 & \texttt{mix75} & 8.03 & 7.49 & 13.89 & 11.15 & 0.44 & 0.50 \\
\midrule
0.6 & WT2 & 13.65 & 7.88 & 54.06 & 16.08 & 0.36 & 0.45 \\
 & C4 & 40.49 & 20.67 & 25.17 & 14.61 & 0.38 & 0.45 \\
 & \texttt{mix75} & 14.27 & 8.40 & 35.03 & 14.77 & 0.40 & 0.46 \\
\midrule
0.4 & WT2 & 63.95 & 12.19 & 353.48 & 41.36 & 0.31 & 0.35 \\
 & C4 & 306.60 & 67.50 & 149.04 & 34.87 & 0.30 & 0.36 \\
 & \texttt{mix75} & 63.25 & 13.48 & 218.52 & 32.24 & 0.31 & 0.38 \\
\bottomrule
\end{tabular*}
\end{table}

\leavevmode{Perplexity is more domain-sensitive. At $r=0.4$, switching LLift's
calibration from WT2 to C4 increases WT2 perplexity from $12.19$ to
$67.50$ while reducing C4 perplexity from $41.36$ to $34.87$.
At this budget, LLift lowers both perplexities relative to the original
checkpoint under all three matched calibration settings.}

\subsection{Low-Rank Prehab and Matched Training Exposure}
\label{app:prehab_relation}

\leavevmode{We compare our Low-Rank Prehab~\citep{qin2025lowrankprehabpreparingneural}
reimplementation with LLift on LLaMA-7B using the same first 256 WT2
training chunks of length 1024 for three epochs ($0.79$M token positions),
with global batch size 16, 48 optimizer steps, and seed 42.
Prehab minimizes CE plus
$\lambda\sum_\ell\|W_\ell T_\ell\|_*^2/\|W_\ell T_\ell\|_F^2$,
using frozen original-model whitening factors, $\lambda=10^{-3}$,
and AdamW with learning rate $10^{-5}$.
LLift retains its default KL objective, $\rho=0.6$ anchor, and $10^{-4}$
learning rate. We apply SVD-LLM to each checkpoint with shared frozen
\texttt{mix75} transforms, identical deployment ranks, and no further
fine-tuning. Table~\ref{tab:prehab_exposure} includes a CE-only control
($\lambda=0$) and default LLift ($11.53$M positions) as references.}

\begin{table}[!htbp]
\centering\small
\setlength{\tabcolsep}{5pt}
\setlength{\belowcaptionskip}{5pt}
{\caption{Training-exposure comparison on LLaMA-7B with SVD-LLM and frozen
\texttt{mix75} calibration. Exposure counts training token positions in
millions, including repeated epochs. Dense denotes the uncompressed
checkpoint; Avg. is seven-task mean plain accuracy.}
\label{tab:prehab_exposure}}
\begin{tabular*}{\linewidth}{@{\extracolsep{\fill}}llrrrr@{}}
\toprule
Budget & Model & Exposure (M) & WT2 PPL $\downarrow$ & C4 PPL $\downarrow$ & Avg. $\uparrow$ \\
\midrule
Dense & Original & --- & 5.68 & 7.34 & 0.54 \\
 & CE fine-tuning & 0.79 & 5.27 & 7.72 & 0.54 \\
 & Prehab (reimpl.) & 0.79 & 5.22 & 7.82 & 0.35 \\
 & LLift & 0.79 & 8.04 & 8.13 & 0.54 \\
 & LLift & 11.53 & 7.38 & 8.23 & 0.54 \\
\midrule
$r=0.8$ & Original & --- & 8.03 & 13.89 & 0.44 \\
 & CE fine-tuning & 0.79 & 7.53 & 15.27 & 0.33 \\
 & Prehab (reimpl.) & 0.79 & 8.04 & 17.74 & 0.33 \\
 & LLift & 0.79 & 9.46 & 11.89 & 0.50 \\
 & LLift & 11.53 & 7.49 & 11.15 & 0.50 \\
\midrule
$r=0.6$ & Original & --- & 14.27 & 35.03 & 0.40 \\
 & CE fine-tuning & 0.79 & 13.44 & 42.18 & 0.39 \\
 & Prehab (reimpl.) & 0.79 & 16.76 & 62.88 & 0.30 \\
 & LLift & 0.79 & 11.73 & 16.63 & 0.46 \\
 & LLift & 11.53 & 8.40 & 14.77 & 0.46 \\
\midrule
$r=0.4$ & Original & --- & 63.25 & 218.52 & 0.31 \\
 & CE fine-tuning & 0.79 & 65.40 & 303.82 & 0.30 \\
 & Prehab (reimpl.) & 0.79 & 94.64 & 459.60 & 0.30 \\
 & LLift & 0.79 & 20.98 & 39.36 & 0.37 \\
 & LLift & 11.53 & 13.48 & 32.24 & 0.38 \\
\bottomrule
\end{tabular*}
\end{table}

\begin{samepage}
\leavevmode{At $r=0.8/0.6/0.4$, LLift at $0.79$M positions is
$0.49/0.14/0.79$ percentage points below default LLift in mean accuracy
(computed from unrounded scores), with higher WT2 and C4 perplexity.
It exceeds our Prehab reimplementation's mean accuracy at all three
budgets. However, the latter's uncompressed accuracy already falls from
$0.54$ to $0.35$. We therefore treat this single-seed comparison as a
reproduction diagnostic, without claiming superiority over a validated
Prehab implementation.}
\par
\end{samepage}

\subsection{Comparison with Decomposable-Net}
\label{app:decomposable_net}

\begin{table}[!htbp]
\centering\small
\setlength{\tabcolsep}{4pt}
\setlength{\belowcaptionskip}{5pt}
{\caption{Comparison with Decomposable-Net on LLaMA-7B at matched training
exposure. D-Net denotes native truncated SVD with global rank allocation.
SVD-LLM denotes activation-whitened SVD with shared calibration and
deployment ranks across checkpoints.
WT2 and C4 report perplexity ($\downarrow$), and Avg. is
seven-task mean plain accuracy ($\uparrow$).}
\label{tab:decomposable_net}}
\begin{tabular*}{\linewidth}{@{\extracolsep{\fill}}lllrrr@{}}
\toprule
Budget & Preparation & Deployment & WT2 PPL $\downarrow$ & C4 PPL $\downarrow$ & Avg. $\uparrow$ \\
\midrule
Dense & Original & --- & 5.68 & 7.34 & 0.54 \\
 & Decomposable-Net & --- & 9.44 & 23.14 & 0.51 \\
 & LLift & --- & 7.38 & 8.23 & 0.54 \\
\midrule
$r=0.8$ & Original & D-Net & 9682.78 & 10689.00 & 0.30 \\
 & Original & SVD-LLM & 8.03 & 13.89 & 0.44 \\
 & Decomposable-Net & D-Net & 9.06 & 33.83 & 0.45 \\
 & Decomposable-Net & SVD-LLM & 9.87 & 43.27 & 0.47 \\
 & LLift & SVD-LLM & 7.49 & 11.15 & 0.50 \\
\midrule
$r=0.6$ & Original & D-Net & 14968.86 & 29171.23 & 0.30 \\
 & Original & SVD-LLM & 14.27 & 35.03 & 0.40 \\
 & Decomposable-Net & D-Net & 11.83 & 60.26 & 0.39 \\
 & Decomposable-Net & SVD-LLM & 12.68 & 85.71 & 0.42 \\
 & LLift & SVD-LLM & 8.40 & 14.77 & 0.46 \\
\midrule
$r=0.4$ & Original & D-Net & 31626.26 & 30290.67 & 0.30 \\
 & Original & SVD-LLM & 63.25 & 218.52 & 0.31 \\
 & Decomposable-Net & D-Net & 20.04 & 139.06 & 0.33 \\
 & Decomposable-Net & SVD-LLM & 34.89 & 460.14 & 0.34 \\
 & LLift & SVD-LLM & 13.48 & 32.24 & 0.38 \\
\bottomrule
\end{tabular*}
\end{table}

\leavevmode{We adapt Decomposable-Net~\citep{yaguchi2021decomposable} to pretrained
LLaMA-7B using next-token CE supervision. We retain its joint dense and
low-rank objective,
\begin{equation}
\mathcal{L}_{\mathrm{DNet}}
=0.5\,\mathcal{L}_{\mathrm{CE}}(W)
+0.5\,\mathcal{L}_{\mathrm{CE}}(\widetilde W),
\end{equation}
where $\widetilde W$ is constructed by truncated SVD at each training
step, with gradients propagated through the decomposition.
We sample a parameter-retention ratio uniformly from $[0.4,0.8]$ and
allocate ranks across the 224 target matrices by globally ordering their
singular values. Relative to the original formulation, we use a pretrained
language model, AdamW, and sampling over parameter-retention ratios.
Training follows LLift's WT2 schedule with sequence length 1024, global
batch size 64, four epochs (176 optimizer steps), and approximately
$11.5$M token positions. We use a learning rate of $10^{-4}$ and seed 42.

We evaluate the Decomposable-Net checkpoint using both its native
truncated-SVD deployment with global rank allocation, denoted D-Net in
Table~\ref{tab:decomposable_net}, and SVD-LLM. The latter uses the same
frozen original-model \texttt{mix75} calibration transforms and deployment
ranks as LLift. Dense denotes each uncompressed checkpoint.}

\leavevmode{Table~\ref{tab:decomposable_net} shows that Decomposable-Net preparation
improves compressed-model accuracy under its native deployment procedure.
LLift achieves higher mean accuracy and lower WT2/C4 perplexity at all
three retention ratios than either Decomposable-Net deployment variant.}

\FloatBarrier
\section{Computational Cost and Resource Usage}
\label{app:cost_records}
For deployment ratios $\mathcal R$, we account for preparation,
calibration, decomposition, and per-budget factor materialization:
\begin{equation}
C_{\mathrm{LLift}}(\mathcal R)=C_{\mathrm{prep}}+C_{\mathrm{cal}}+
C_{\mathrm{SVD}}+\sum_{r\in\mathcal R}C_{\mathrm{materialize}}(r).
\label{eq:deployment_cost}
\end{equation}
With one prepared model and fixed activation transforms, we apply weighted
SVD to each target matrix of the prepared checkpoint once and materialize
factors at each budget. We compute GPU-hours as elapsed stage time times
the number of devices held, counting shared work once and excluding
quality evaluation.

\subsection{Preparation and Training Resources}
\label{app:preparation_resources}

\leavevmode{Table~\ref{tab:v3_resources} reports preparation costs for all five models
on A100-80GB GPUs, covering WT2 whitening, basis construction,
optimization, and merging $W^\star=Q_0B^\star+R_0$.
We compute stage costs from recorded runtimes and log-based estimates,
including loading and saving only where timed. Deployment costs are
accounted for separately in Eq.~\ref{eq:deployment_cost}.}

\begin{table}[!htbp]
\centering\small
\setlength{\tabcolsep}{4pt}
\setlength{\belowcaptionskip}{5pt}
{\caption{LLift preparation-stage costs in A100-80GB GPU-hours.
Subtotals are computed before rounding.}
\label{tab:v3_resources}}
\begin{tabular*}{\linewidth}{@{\extracolsep{\fill}}lrrrrr@{}}
\toprule
Model & Whitening & Bases & Optimization & Merge & Subtotal \\
\midrule
LLaMA-7B & 0.40 & 0.09 & 1.56 & 0.02 & 2.07 \\
Qwen3-8B & 0.55 & 0.10 & 1.75 & 0.01 & 2.42 \\
Gemma-2-9B & 0.87 & 0.15 & 1.43 & 0.01 & 2.47 \\
LLaMA-13B & 3.97 & 0.15 & 4.97 & 0.02 & 9.11 \\
LLaMA-30B & 11.48 & 0.55 & 8.17 & 0.05 & 20.25 \\
\bottomrule
\end{tabular*}
\end{table}

\leavevmode{Optimization uses four GPUs, while basis
construction and merging use one. Whitening uses one GPU for the 7B/8B/9B
models and holds four for LLaMA-13B/30B, including during single-GPU transform
derivation.}

\leavevmode{On four A100-80GB GPUs, FlexRank (Table~\ref{tab:elastic}) trains on about $12$M
FineWeb-Edu token positions and takes $27.0$ GPU-hours, of which $24.0$ are
per-layer profiling for its rank search. LLift's preparation takes $2.07$
GPU-hours with about $11.5$M WT2 token positions. Because the corpora differ,
these costs describe each method's own workflow rather than a matched comparison.}

\begin{table}[!htbp]
\centering\small
\setlength{\tabcolsep}{4pt}
\setlength{\belowcaptionskip}{5pt}
{\caption{LLift optimization resources on four A100-80GB GPUs.
Peak denotes allocated GPU memory on rank 0.}
\label{tab:training_resources}}
\begin{tabular*}{\linewidth}{@{\extracolsep{\fill}}lrrrrc@{}}
\toprule
Model & Steps & Time (s) & Positions/s/GPU & Peak (GiB) & CPU offload \\
\midrule
LLaMA-7B & 176 & 1405.4 & 2045 & 32.18 & No \\
Qwen3-8B & 156 & 1573.2 & 1601 & 47.74 & No \\
Gemma-2-9B & 152 & 1289.1 & 1887 & 60.67 & No \\
LLaMA-13B & 176 & 4472.4 & 643 & 42.89 & Yes \\
LLaMA-30B & 176 & 7350.2 & 391 & 41.12 & Yes \\
\bottomrule
\end{tabular*}
\end{table}

\leavevmode{Table~\ref{tab:training_resources} reports optimization wall-clock time,
throughput, and peak allocated GPU memory. All models use BF16, ZeRO-2,
sequence length 1024, and global batch size 64 on four A100-80GB GPUs.
The 7B/8B/9B models use microbatch size four and four accumulation steps
per GPU. LLaMA-13B/30B use CPU optimizer offload, microbatch size one,
and 16 accumulation steps. We estimate token positions per second per GPU
as sample throughput $\times1024/4$, including repeated epochs and padding.
Peak memory covers rank 0 during optimization and excludes host memory.}

\subsection{Budget-Specific Fine-Tuning and Deployment Timings}
\label{app:workflow_costs}

We follow SVD-LLM's sequential LoRA procedure, adapting $U$ and then
$V$ on Alpaca for each budget (Appendix~\ref{app:budget_finetuning}).
This takes approximately $5.4$ GPU-hours per budget on average, totaling $16.1$
GPU-hours for $r\in\{0.8,0.6,0.4\}$, excluding compression and
deployment.

\leavevmode{Table~\ref{tab:matched_recovery_costs} reports estimated pipeline costs for
the matched LLaMA-7B recovery comparison in
Appendix~\ref{app:matched_recovery}. Pipeline costs include preparation
or recovery, deployment calibration, and SVD compression.
LLift requires $3.13$ A100 GPU-hours for three budgets,
compared with $5.10$ when each recovery run receives LLift's training
exposure. Training positions are summed across runs.
We sum stage-level GPU-hours, count shared work once, and
exclude evaluation. The $2.07$ GPU-hours in Table~\ref{tab:v3_resources}
cover preparation alone. LLift supports additional deployment budgets without
further training.}

\begin{table}[!htbp]
\centering\small{}
\setlength{\tabcolsep}{4pt}
\setlength{\belowcaptionskip}{5pt}
{\caption{LLaMA-7B training and pipeline costs for three
deployment budgets ($r=0.8/0.6/0.4$), in A100 GPU-hours.}
\label{tab:matched_recovery_costs}}
\begin{tabular*}{\linewidth}{@{\extracolsep{\fill}}lrrrr@{}}
\toprule
Method & Training runs & Positions (M) & Optimization & Pipeline \\
\midrule
LLift & 1 & 11.50 & 1.55 & 3.13 \\
Recovery & 3 & 34.50 & 4.00 & 5.10 \\
\bottomrule
\end{tabular*}
\end{table}

\leavevmode{LLift's optimization time comes from a profile with matching
factor shapes and training steps. Both methods use the decomposition
cost recorded for a four-budget batched SVD run.}

On A100 GPUs, one-shot ZS-SVD and ZS-SVD $5\times$ take $0.25$ and
$2.11$ GPU-hours, respectively, for one deployment budget, excluding
LLift preparation.

\FloatBarrier
\subsection{Inference Resources}
\label{app:inference_resources}

\leavevmode{We benchmark dense and factorized shapes with random FP16 weights on
one A100-80GB GPU using eager Hugging Face execution with SDPA
and no compilation. The factorized shapes use the SVD-LLM rank rule
and two-factor layout shared by LLift, with parameter counts matching
all 20 model--budget configurations. This benchmark measures the
corresponding structures, without timing the trained compressed checkpoints.

We measure prefill latency for 2048 tokens at batch size one. Decode
generates 128 tokens from a four-token prompt with KV caching and
greedy decoding at batch sizes 1, 16, and 64. Each setting uses three
warmup iterations and ten timed iterations.
Table~\ref{tab:inference_resources} reports weight memory, peak
allocated GPU memory during decoding at batch size 64, prefill latency,
and aggregate decode throughput.}

\begin{table}[!htbp]
\centering\small
\setlength{\tabcolsep}{3pt}
\setlength{\belowcaptionskip}{5pt}
{\caption{Inference resources for dense and factorized shapes with random
FP16 weights on one A100-80GB GPU. Peak denotes allocated GPU memory
during decoding at batch size 64.}
\label{tab:inference_resources}}
\begin{tabular*}{\linewidth}{@{\extracolsep{\fill}}lcrrrrrr@{}}
\toprule
\multirow{2}{*}{Model} & \multirow{2}{*}{Budget} &
\multicolumn{2}{c}{Memory (GiB)} & \multirow{2}{*}{Prefill (ms)} &
\multicolumn{3}{c}{Decode (tokens/s)} \\
\cmidrule(lr){3-4}\cmidrule(l){6-8}
 & & Weights & Peak & & $B=1$ & $B=16$ & $B=64$ \\
\midrule
LLaMA-7B & Dense & 12.55 & 16.73 & 157 & 40.4 & 624.8 & 2505.5 \\
 & 0.8 & 10.14 & 14.32 & 210 & 31.7 & 456.2 & 1831.6 \\
 & 0.6 & 7.72 & 11.91 & 264 & 31.3 & 459.9 & 1728.4 \\
 & 0.4 & 5.31 & 9.50 & 173 & 30.6 & 459.8 & 1860.1 \\
\midrule
Qwen3-8B & Dense & 15.26 & 16.62 & 183 & 26.5 & 419.0 & 1689.2 \\
 & 0.8 & 12.67 & 14.03 & 412 & 23.0 & 339.7 & 1332.2 \\
 & 0.6 & 10.08 & 11.44 & 336 & 22.8 & 333.3 & 1332.0 \\
 & 0.4 & 7.49 & 8.85 & 195 & 22.7 & 341.9 & 1353.1 \\
\midrule
Gemma-2-9B & Dense & 17.21 & 20.25 & 235 & 22.1 & 353.4 & 1410.1 \\
 & 0.8 & 14.11 & 17.14 & 513 & 18.3 & 278.4 & 1004.0 \\
 & 0.6 & 11.01 & 14.04 & 227 & 18.3 & 271.8 & 1116.8 \\
 & 0.4 & 7.91 & 10.94 & 216 & 18.8 & 275.8 & 1099.9 \\
\midrule
LLaMA-13B & Dense & 24.24 & 30.75 & 274 & 27.7 & 407.2 & 1646.2 \\
 & 0.8 & 19.51 & 26.02 & 331 & 21.0 & 312.5 & 1240.1 \\
 & 0.6 & 14.79 & 21.29 & 427 & 22.1 & 336.5 & 1120.9 \\
 & 0.4 & 10.06 & 16.57 & 203 & 21.1 & 325.7 & 1309.3 \\
\midrule
LLaMA-30B & Dense & 60.59 & 73.19 & 665 & 17.8 & 261.0 & 836.2 \\
 & 0.8 & 48.62 & 61.23 & 910 & 14.8 & 210.2 & 639.4 \\
 & 0.6 & 36.67 & 49.27 & 558 & 14.7 & 224.4 & 777.7 \\
 & 0.4 & 24.71 & 37.31 & 993 & 13.9 & 200.2 & 598.1 \\
\bottomrule
\end{tabular*}
\end{table}

\leavevmode{Dense-to-compressed weight-memory ratios range from $1.20$ to $2.45$.
Decode throughput is $0.68$--$0.93$ times the dense baseline, while
prefill latency varies with model and budget. On the shared node, repeated
LLaMA-7B runs differ by up to approximately $18\%$ in decode throughput.
Timing equivalence with trained compressed checkpoints remains unverified.}

\section{GSM8K Generation}
\label{app:gsm8k}

\leavevmode{We evaluate Qwen3-8B on GSM8K~\citep{cobbe2021trainingverifierssolvemath}
to assess multi-step generation. We use the \texttt{gsm8k} task in
lm-eval 0.4.11 on all 1,319 test examples, with five-shot prompting,
greedy decoding, and seed 42. All models use the same prompts without
a chat template and run in FP16 with vLLM 0.17.1 and a maximum total
sequence length of 2048 tokens. We reconstruct each compressed weight
as the product of its two factors before evaluation.
Table~\ref{tab:gsm8k_qwen3} reports strict exact-match accuracy.
The original dense model achieves $88.63\%$ accuracy.
LLift outperforms SVD-LLM and ZS-SVD at $r=0.8/0.6$.
All methods approach zero at $r=0.4$. These results show gains
in compressed-model generation, although substantial degradation remains.}

\begin{table}[!htbp]
\small
\setlength{\tabcolsep}{12pt}
{\caption{Qwen3-8B GSM8K strict exact-match accuracy (\%, $\uparrow$).}
\label{tab:gsm8k_qwen3}}
\begin{center}
\begin{tabular}{@{}lrrr@{}}
\toprule
Method & $r=0.8$ & $r=0.6$ & $r=0.4$ \\
\midrule
SVD-LLM & 24.94 & 0.00 & 0.00 \\
ZS-SVD & 13.19 & 0.00 & 0.00 \\
LLift & 64.67 & 17.13 & 0.38 \\
\bottomrule
\end{tabular}
\end{center}
\end{table}

\end{document}